\documentclass[10pt,twocolumn,letterpaper]{article}

\usepackage{cvpr}              % To produce the CAMERA-READY version
\usepackage{multirow}
\usepackage{url}
\usepackage{epsfig}
\usepackage{graphicx} 
\usepackage{amsmath}
\usepackage{amssymb}
\usepackage{adjustbox}
\usepackage{multirow}
\usepackage{makecell}
\usepackage{multicol}
\usepackage{mathrsfs}
\usepackage[ruled]{algorithm2e}
\usepackage{cellspace}

\usepackage{makecell}
\usepackage{bm} 
\usepackage{booktabs}
\usepackage{bbding}
\usepackage{mdframed}
\usepackage[accsupp]{axessibility} % Improves PDF readability for those with visual impairments.
\usepackage[dvipsnames]{xcolor}

\newcommand{\hide}[1]{}

\definecolor{cvprblue}{rgb}{0.21,0.49,0.74}
\usepackage[pagebackref,breaklinks,colorlinks,citecolor=cvprblue]{hyperref}

\def\paperID{14567} % *** Enter the Paper ID here
\def\confName{CVPR}
\def\confYear{2024}

\title{CogAgent: A Visual Language Model for GUI Agents}

\author{Wenyi Hong\textsuperscript{1}\thanks{Work was done when interned at Zhipu AI.} \ \ Weihan Wang\textsuperscript{1}\footnotemark[1]\ \ Qingsong Lv\textsuperscript{2} \ \ Jiazheng Xu\textsuperscript{1}\footnotemark[1] \ \ Wenmeng Yu\textsuperscript{2} \ \ \\ 
Junhui Ji\textsuperscript{2} \ \ 
Yan Wang\textsuperscript{2} \ \ Zihan Wang\textsuperscript{1}\footnotemark[1] \ \  Yuxuan Zhang\textsuperscript{2}\footnotemark[1] \ \ Juanzi Li\textsuperscript{1} \ \ \\
 Bin Xu\textsuperscript{1} \ \ Yuxiao Dong\textsuperscript{1} \ \ Ming Ding\textsuperscript{2}\textsuperscript{$\dagger$} \ \ Jie Tang\textsuperscript{1}%\textsuperscript{$\ddagger$}
\thanks{Corresponding authors}
\\
  \textsuperscript{1}Tsinghua University\ \ \textsuperscript{2}Zhipu AI\ \ %\textsuperscript{$\ddagger$}corresponding authors
  \\
  \texttt{\{hwy22@mails, jietang@\}.tsinghua.edu.cn, ming.ding@zhipuai.cn} \\
  }

\definecolor{dt}{gray}{0.6}
\definecolor{dtdark}{gray}{0.5}

\definecolor{MyBlue}{HTML}{C6E9FF}

\begin{document}

\maketitle

 \begin{abstract}

People are spending an enormous amount of time on digital devices through graphical user interfaces (GUIs), e.g., computer or smartphone screens.  
%In modern life, a significant amount of people's time is spent interacting with computers or smartphones. 
Large language models (LLMs) such as ChatGPT can assist people in tasks like writing emails, but struggle to understand and interact with GUIs, thus limiting their potential to increase automation levels. 
In this paper, we introduce CogAgent, an 18-billion-parameter visual language model (VLM) specializing in GUI understanding and navigation. 
By utilizing both low-resolution and high-resolution image encoders, CogAgent supports input at a resolution of $1120\times 1120$, enabling it to recognize tiny page elements and text.
As a generalist visual language model, CogAgent achieves the state of the art on five text-rich and four general VQA benchmarks, including VQAv2, OK-VQA, Text-VQA, ST-VQA, ChartQA, infoVQA, DocVQA, MM-Vet, and POPE. 
CogAgent, using only screenshots as input, outperforms LLM-based methods that consume extracted HTML text on both PC and Android GUI navigation tasks---Mind2Web and AITW, advancing the state of the art. 
The model and codes are available at \url{https://github.com/THUDM/CogVLM}, with a new version of CogAgent-9B-20241220 available at \url{https://github.com/THUDM/CogAgent}.
\end{abstract}

\hide{
In modern life, a significant amount of people's time is spent interacting with computers or smartphones. Current large language models can assist people in tasks like writing emails, but they struggle to understand and operate GUIs, thus limiting their ability to achieve higher levels of automation. In this paper, we introduce CogAgent, an 18-billion-parameter visual language model specializing in GUI understanding and navigation. With a novel design utilizing both low-resolution and high-resolution image encoders, CogAgent supports input at a resolution of $1120\times 1120$, enabling it to recognize tiny page elements and text.
As a generalist visual language model, CogAgent achieves state-of-the-art on 5 text-rich and 4 general VQA benchmarks, including VQAv2, OK-VQA, Text-VQA, ST-VQA, ChartQA, infoVQA, DocVQA, MM-Vet and POPE. CogAgent, with only screenshot as input, for the first time surpasses LLM-based methods with extracted HTML input on both PC and Android GUI navigation tasks, Mind2Web and AITW, achieving new state-of-the-art performance. The model and codes are available at \url{https://github.com/THUDM/CogVLM}.
}    
 \section{Introduction}

\begin{figure*}[t]
    \centering
    \includegraphics[width=0.99\linewidth]{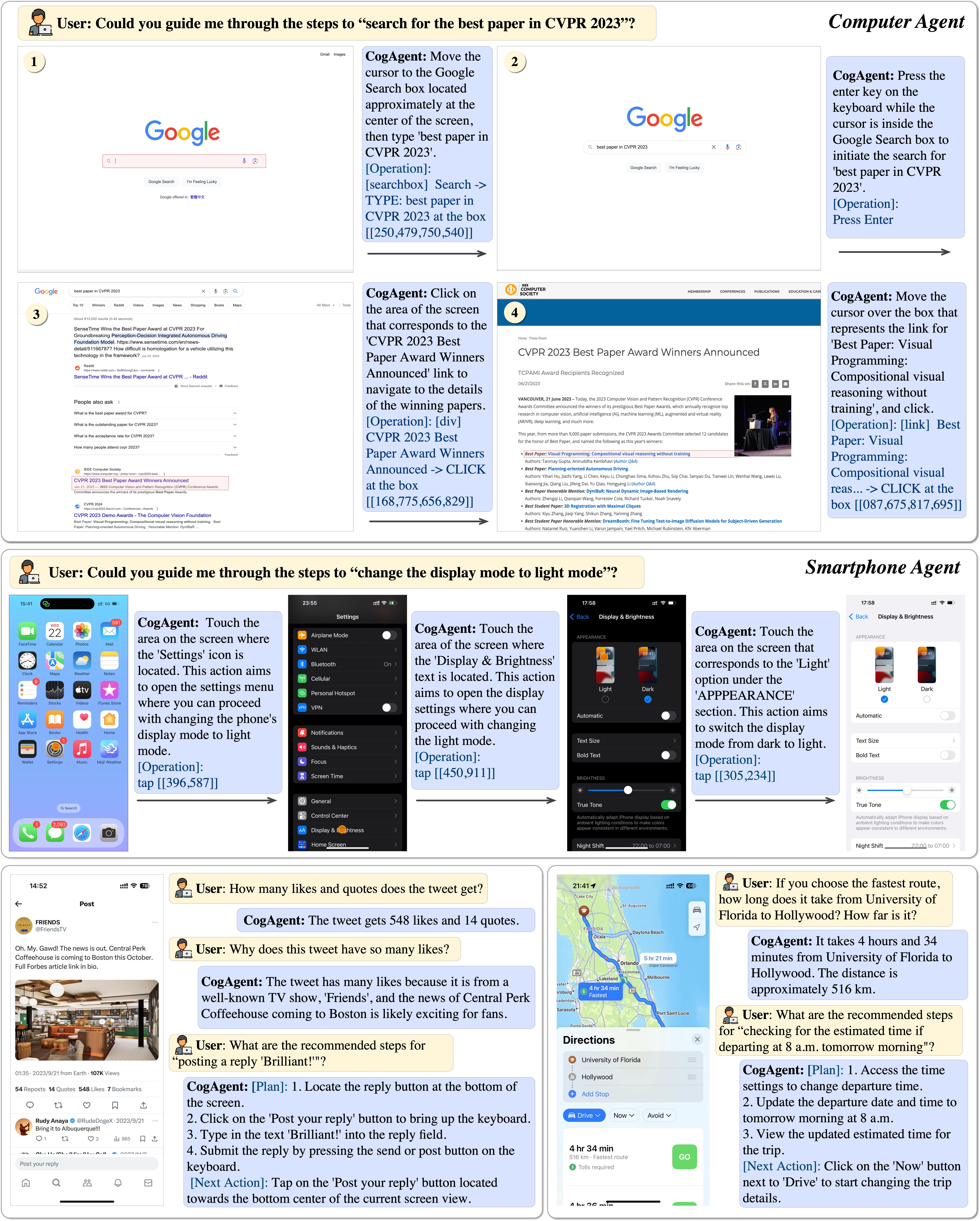}
    \caption{Samples of visual agents generated by CogAgent. More samples are demonstrated in the Appendix. }
    \label{fig:demo}
\end{figure*}

Autonomous agents in the digital world are ideal assistants that many modern people dream of. 
% Imagine this: typing in a task description and being able to sit back, sip coffee, and watch as tasks like online ticket booking, web searching, file operations, and PowerPoint creation are automatically completed. 
Picture this scenario: You type in a task description, then relax and enjoy a cup of coffee while watching tasks like booking tickets online, conducting web searches, managing files, and creating PowerPoint presentations get completed automatically.

Recently, the emergence of agents based on large language models (LLMs) is bringing us closer to this dream. For example, AutoGPT~\cite{autogpt}, a 150,000-star open-source project, leverages ChatGPT~\cite{openai2022chatgpt} to integrate language understanding with pre-defined actions like Google searches and local file operations.
Researchers are also starting to develop agent-oriented LLMs~\cite{zeng2023agenttuning,chen2023fireact}.
However, the potential of purely language-based agents is quite limited in real-world scenarios, as most applications interact with humans through Graphical User Interfaces (GUIs), which are characterized by the following perspectives:
\begin{itemize}
    % \item The two-dimensional spatial structure of graphical information and page elements contains important information that is difficult to convey in words.
    % \item In many cases, there are no standard APIs for interaction.
    \item Standard APIs for interaction are often lacking.
    % \item Important information including text in images, icons and spatial relations are difficult to directly convey in words.
    \item Important information including icons, images, diagrams, and spatial relations are difficult to directly convey in words.
    % \item Even for GUIs with textual representation (e.g. View Hierarchy in Android), uncaptured elements (e.g. WebViews and Canvas) make it difficult to grasp the page's functionality through text alone.
    \item Even in text-rendered GUIs like web pages, elements like canvas and iframe cannot be parsed to grasp their functionality via HTML.
    % \item Even for web pages with corresponding HTML, reliance on JavaScript for dynamic loading makes it hard to understand the page's functionality through text alone.
\end{itemize}
% For LLM-based agents, the task cannot be proceeded if the path of solution contains interaction with any GUI.
%, greatly limiting the scope of applications.

Agents based on visual language models (VLMs) have the potential to overcome these limitations. Instead of relying exclusively on textual inputs such as  HTML~\cite{nakano2021webgpt} or OCR results~\cite{rawles2023android}, VLM-based agents directly perceive visual GUI signals. Since GUIs are designed for human users, VLM-based agents can perform as effectively as humans, as long as the VLMs match human-level vision understanding. In addition, VLMs are also capable of skills such as extremely fast reading and programming that are usually beyond the reach of most human users, extending the potential of VLM-based agents.
% Previous attempts adopt visual features as auxiliary only when dealing with scenarios, e.g. WebShop~\cite{yao2022webshop}, which require object recognition.
A few prior studies utilized visual features merely as auxiliaries in specific scenarios. e.g. WebShop~\cite{yao2022webshop} which employs visual features primarily for object recognition purposes.
% With the rapid development of VLM, can we now rely solely on visual inputs rather than HTML to achieve universality on GUIs?
With the rapid development of VLM, can we naturally achieve universality on GUIs by relying solely on visual inputs?

In this work, we present CogAgent, a visual language foundation model specializing in GUI understanding and planning while maintaining a strong ability for general cross-modality tasks. 
By building upon CogVLM~\cite{wang2023cogvlm}---a recent open-source VLM, CogAgent tackles the following challenges for building GUI agents:
\begin{itemize}
    \item \textbf{Training Data.} Most current VLMs are pre-trained on datasets like LAION~\cite{schuhmann2022laion}, consisting of natural images on the Web. However, we notice that the GUI images share a different distribution from natural images. 
    We thus construct a large-scale annotated dataset about GUIs and OCR for continual pre-training.

    \item \textbf{High-Resolution vs. Compute.} In GUIs, tiny icons and text are ubiquitous, and it is hard to recognize them in commonly-used $224\times224$ resolution. 
    However, increasing the resolution of input images results in significantly long sequence length in language models. 
    For example, a $1120\times 1120$ image corresponds to a sequence of $6400$ tokens if the patch size is $14$, demanding excessive training and inference compute. 
    To address this, we design a cross-attention branch that allows for a trade-off between the resolution and the hidden size within a proper computation budget. 
    Specifically, we propose to combine the original large ViT~\cite{dosovitskiy2020image} (4.4B parameters) used in CogVLM~\cite{wang2023cogvlm} and a new small \emph{high-resolution cross-module} (with image encoder of 0.30B parameters) to jointly model visual features.
\end{itemize}

\noindent Our experiments show that:
\begin{itemize}
    \item CogAgent tops popular GUI understanding and decision-making benchmarks, including AITW~\cite{rawles2023android}
    and Mind2Web~\cite{deng2023mind2web}. 
    To the best of our knowledge, this is the first time that a generalist VLM can outperform LLM-based methods with extracted structured text.

    \item Though CogAgent focuses on GUIs, it achieves state-of-the-art generalist performance on nine visual question-answering benchmarks including VQAv2~\cite{antol2015vqa}, OK-VQA~\cite{marino2019ok}, TextVQA~\cite{singh2019towards}, ST-VQA~\cite{biten2019scene}, ChartQA~\cite{masry2022chartqa}, infoVQA~\cite{mathew2022infographicvqa}, DocVQA~\cite{mathew2021docvqa}, MM-Vet~\cite{yu2023mm}, and POPE~\cite{li2023evaluating}.

    \item The separated design of high- and low-resolution branches in CogAgent significantly lows the compute cost for consuming high-resolution images, e.g., the number of the floating-point operations (FLOPs) for CogAgent-18B with $1120 \times 1120$ inputs is less than half that of CogVLM-17B with its default $490\times 490$ inputs. 
    
    \end{itemize}

CogAgent is open-sourced at \url{https://github.com/THUDM/CogVLM}, with a new version of CogAgent-9B-20241220 available at \url{https://github.com/THUDM/CogAgent}. 
It represents an effort to promote the
future research and application of AI agents, facilitated by advanced VLMs.

%%% end of hide

 \section{Method}

In this section, we will first introduce the architecture of CogAgent, especially the novel high-resolution cross-module, and then illustrate the process of pre-training and alignment in detail. 

\subsection{Architecture}
\begin{figure}
    \centering
    \includegraphics[width=\linewidth]{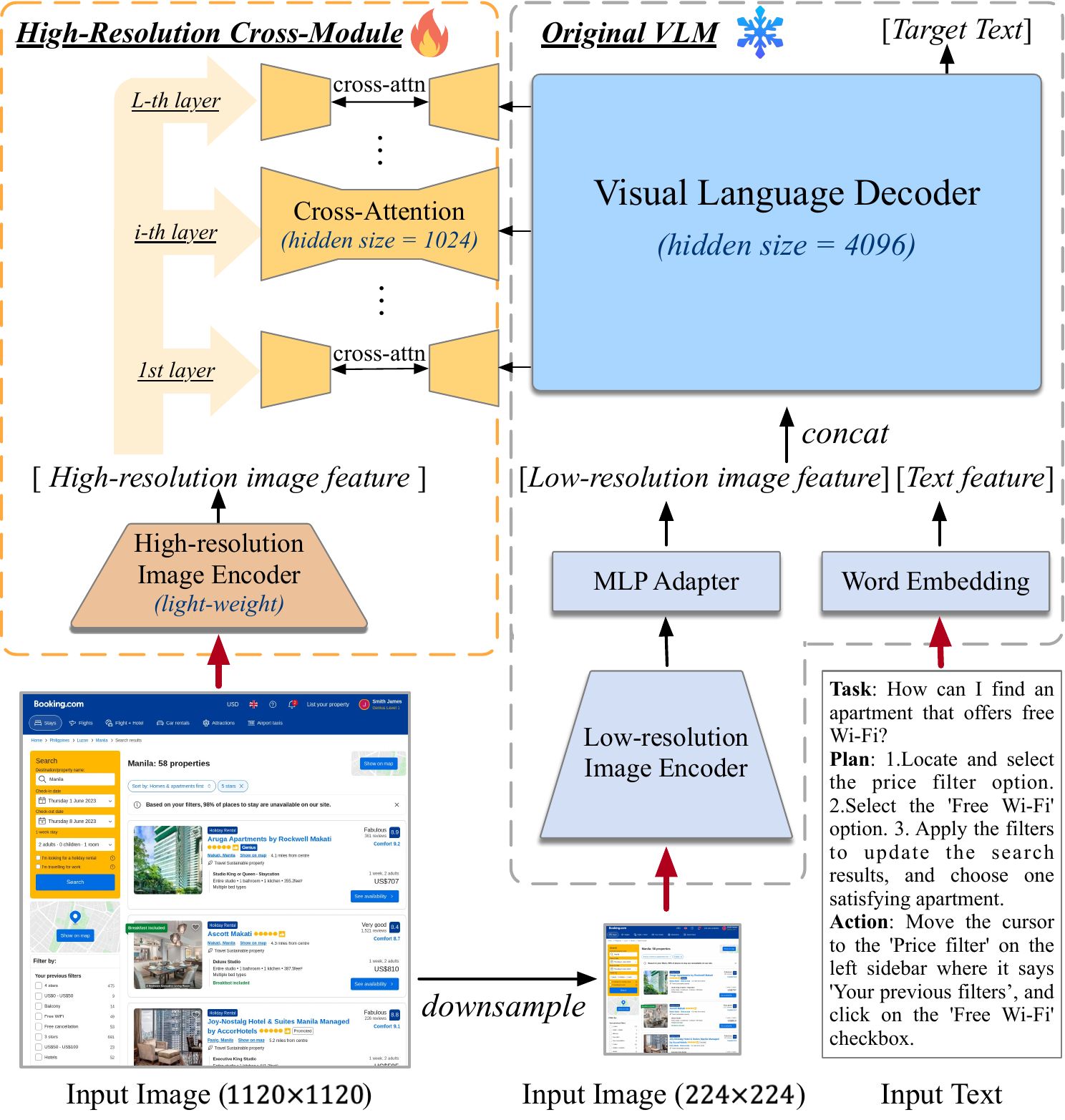}
    \caption{Model architecture of CogAgent. We adopt CogVLM as the original VLM. }
    \label{fig:arch}
    \vspace{-0.5cm}
\end{figure}

The architecture of CogAgent is depicted in \cref{fig:arch}. We build our model based on a pre-trained VLM (on the right side of the image), and propose to add a cross-attention module to process high-resolution input (on the left side of the image). As our base VLM, We select CogVLM-17B~\cite{wang2023cogvlm}, an open-sourced and state-of-the-art large vison-language model. Specifically, We employ EVA2-CLIP-E~\cite{sun2023eva} as the encoder for low-resolution images (224$\times$224 pixels), complemented by an MLP adapter that maps its output into the feature space of the visual-language decoder. The decoder, a pre-trained language model, is enhanced with a visual expert module introduced by ~\citet{wang2023cogvlm} to facilitate a deep fusion of visual and language features. The decoder processes a combined input of the low-resolution image feature sequence and text feature sequence, and autoregressively outputs the target text.

Similar to most VLMs, the original CogVLM can only accommodate images of relatively low resolution (224 or 490), which hardly meets the demands of GUI where the screen resolution of computers or smartphones is typically 720p ($1280\times720$ pixels) or higher. It is a common problem among VLMs, e.g. LLaVA~\cite{liu2023visual} and PALI-X~\cite{chen2023pali} are pre-trained at a low resolution of $224\times224$ on the general domain. 
The primary reason is that high-resolution image brings prohibitive time and memory overhead: VLMs usually concatenate text and image feature sequence as input to the decoder, thus the overhead of self-attention module is quadratic to the number of visual tokens (patches), which is quadratic to the image's side length.
There are some initial attempts to reduce costs for high-resolution images. For instance, Qwen-VL~\cite{bai2023qwen} proposes a position-aware vision-language adapter to compress image features, but only reduces sequence length by four and has a maximum resolution of $448\times448$. Kosmos-2.5~\cite{lv2023kosmos} adopts a Perceiver Resampler module to reduce the length of the image sequence. However, the resampled sequence is still long for self-attention in the large visual-language decoder (2,048 tokens), and can only be applied to restricted text recognition tasks. 
% \hwy{flamingo?} qwen-vl

Therefore, we propose a novel \textit{high-resolution cross-module} as a potent complement to the existing structure for enhancing understanding at high resolutions, which not only maintains efficiency confronting high-resolution images, but also offers flexible adaptability to a variety of visual-language model architectures.  

% suffers from prohibitive computational costs 
\subsection{High-Resolution Cross-Module}
The structural design of \textit{high-resolution cross-module} is mainly based on the following observations:
\begin{enumerate}
    \item At a modest resolution such as $224\times224$, images can depict most objects and layouts effectively, yet the resolution falls short in rendering text with clarity. Hence, our new high-resolution module should emphasize text-related features, which are vital for understanding GUIs.
    \item While pre-trained VLMs in general domain often need large hidden sizes (e.g. 4,096 in PALI-X and CogVLM, 5,120 in LLaVA), VLMs tailored for text-centered tasks like document OCR require smaller hidden sizes to achieve satisfying performance (e.g. 1,536 in Kosmos-2.5 and Pix2Struct~\cite{lee2023pix2struct}). This suggests that text-related features can be effectively captured using smaller hidden sizes. 
\end{enumerate}

As shown in \cref{fig:arch}, the high-resolution cross-module acts as a new branch for higher-resolution input, which accepts images of size $1120\times1120$ pixels in our implementation. Different from the original low-resolution input branch, the high-resolution cross-module adopts a much smaller pre-trained vision encoder (visual encoder of EVA2-CLIP-L~\cite{sun2023eva} in our implementation, 0.30B parameters), and uses cross-attention of a small hidden size to fuse high-resolution image features with every layer of VLLM decoder, thus reducing the computational cost. To be concrete, for an input image, it is resized to $1120\times1120$ and $224\times224$ and fed into the high-resolution cross-module and the low-resolution branch respectively, then encoded into image feature sequences $X_{\text{hi}}$ and $X_{\text{lo}}$ with two distinct-sized image encoders in parallel. The visual language decoder retains its original computations, while the only change is to integrate a cross-attention between $X_{\text{hi}}$ and hidden states in every decoder layer. 

Formally, suppose that the input hidden states of the i-th attention layer in the decoder are $X_{\text{in}_i} \in \mathbb{R}^{B\times (L_{I_{\text{lo}}}+L_T) \times D_{\text{dec}}}$, and the output hidden states of cross-module's image encoder are $X_{\text{hi}} \in \mathbb{R}^{B\times (L_{I_{\text{hi}}}) \times D_{\text{hi}}}$, where B is the batch size, $L_{I_{\text{lo}}}$, $L_{I_{\text{hi}}}$ and $L_T$ are the lengths of the low-resolution image, high-resolution image and text sequences, $D_{\text{dec}}$ and $D_{\text{hi}}$ is the hidden size of the decoder and high-resolution encoder's output respectively. 
% \xjz{May need to explain what $X_{i}'$ and $X_{\text{out}_i}$ are since there is no explanation to these two before.} 
Each layer's attention procedure can be formulated as
\begin{align}
    X_{i}' &= \text{MSA}(\text{layernorm}(X_{\text{in}_i})) + X_{\text{in}_i}, \label{msa} \\
    X_{\text{out}_i} &= \text{MCA}(\text{layernorm}(X_{i}'), X_{\text{hi}}) + X_{i}', \label{eq:mca}
\end{align}
where MSA and MCA represent multi-head self-attention with visual expert and multi-head cross-attention, while $X_{i}'$ and $X_{\text{out}_i}$ represent their respective output features with the residual connection. To implement cross-attention between them, we add learnable transformation matrices $W_{K_{\text{cross}}}^i, W_{V_{\text{cross}}}^i \in \mathbb{R}^{D_{\text{hi}} \times D_{\text{cross}}}$ to get $K_{\text{cross}}^i=X_{\text{hi}} W_{K_{\text{cross}}}^i$, $V_{\text{cross}}^i=X_{\text{hi}} W_{V_{\text{cross}}}^i \in \mathbb{R}^{L_{I_{\text{hi}}} \times D_{\text{cross}}} $, and $W_{Q_{\text{cross}}}^i \in \mathbb{R}^{D_{\text{dec}} \times D_{\text{cross}}}$ to get $Q_{\text{cross}}^i=X_i' W_{Q_{\text{cross}}}^i \in \mathbb{R}^{(L_{I_{\text{lo}}}+L_T) \times D_{\text{cross}}}$ in every decoder layer. With the residual connection in Eq.~\ref{eq:mca}, the cross-attention with high-resolution images can be perceived as a complement to the features of low-resolution images, thereby effectively utilizing the previous pre-trained model in low resolution. 

\textbf{Computational complexity.} Let the number of attention head be $H_{\text{cross}}$ and $H_{\text{dec}}$ in cross-attention and self-attention, and the dimension of each head be $d_{\text{cross}} = D_{\text{cross}}/{H_{\text{cross}}}$ and $d_{\text{dec}} = D_{\text{dec}}/{H_{\text{dec}}}$. If using our high-resolution cross-module, the computational complexity of attention is
\begin{equation}
\begin{split}
\text{T}_{\text{improved}} = \mathbf{O}\bigl( &(L_{I_{\text{lo}}} + L_T) L_{I_{\text{hi}}} H_{\text{cross}} d_{\text{cross}} \\
&+ (L_{I_{\text{lo}}} + L_T)^2 H_{\text{dec}} d_{\text{dec}} \bigr).
\end{split}
\end{equation}
 Note that $d_{\text{cross}}$ and $H_{\text{cross}}$ can be flexibly adjusted according to computational budget and model performance. If not utilizing the high-resolution cross-module and directly substituting low-resolution images with high-resolution ones, the computational complexity would be
\begin{align}
\text{T}_{\text{original}} = \mathbf{O}\bigl((L_{I_{\text{hi}}} + L_T)^2 H_{\text{dec}} d_{\text{dec}} \bigr).
\end{align}

In our implementation, $d_{\text{cross}}=32$, $H_{\text{cross}}=32$, and we inherits $d_{\text{dec}}=128$, $H_{\text{dec}}=32$ from CogVLM-17B. Both high- and low-resolution encoders patchify images with $14\times14$-pixel patches, thus $L_{I_{\text{hi}}}=6400$, $L_{I_{\text{lo}}}=256$. Our method leads to at least $\frac{L_{I_{\text{hi}}}+L_{T}}{L_{I_{\text{lo}}}+L_{T}} = \frac{6400+L_{T}}{256+L_{T}} \times$ acceleration which is a stringent lower bound (refer to Appendix for detailed derivation),  and reduces memory overhead at the same time.

\subsection{Pre-training}
To enhance the model's ability to comprehend high-resolution images and adapt it for GUI application scenarios, we focus our pre-training efforts on the following aspects: the capability to recognize texts of various sizes, orientations, and fonts in high-resolution images, the grounding ability of text and objects in the image, and a specialized understanding capability for GUI imagery such as web page. We divide our pre-train data into three parts based on the aforementioned aspects, with samples in the Appendix. All the pre-training data are derived from publicly available datasets. The construction methods are detailed below. 
% \hwy{refer to appendix, show data case; augmentation code}

\noindent \textbf{Text recognition.} Our data includes (1) Synthetic renderings with text from language pre-training dataset (80M). This is similar to the Synthetic Document Generator in ~\citet{kim2022ocr}, with text of varying font, size, color and orientation, and diverse image background from LAION-2B~\cite{schuhmann2022laion}. (2) Optical Character Recognition (OCR) of natural images (18M). We collect natural images from COYO~\cite{kakaobrain2022coyo-700m} and LAION-2B~\cite{schuhmann2022laion} and employ Paddle-OCR~\cite{du2020pp} to extract the texts and their bounding boxes, and filter out images with no text boxes. 
% Paddle-OCR may introduce some errors, which can be ameliorated through integration with other pre-training datasets and subsequent fine-tuning processes. 
(3) Academic documents (9M). We follow Nougat~\cite{blecher2023nougat} to construct image-text pairs including text, formula and tables from the source code (LaTeX) release on arXiv. For (1)(3), we apply the same data augmentation as Nougat.
%  which includes erosion, gaussian noise, gaussian blur, image compression, and elastic transform, etc. 
For (2), we additionally employed more aggressive rotation and flipping data augmentation techniques.
% , thereby enhancing the model's robustness in recognizing text.

\noindent \textbf{Visual grounding.} It is imperative for GUI agents to possess the capability to accurately comprehend and locate diverse elements within images. 
% Therefore, we incorporated a range of grounding data into pre-training. 
We follow CogVLM~\cite{wang2023cogvlm} to use a constructed visual grounding dataset of 40M images with image-caption pairs sampled from LAION-115M~\cite{li2023blip}, which associate entities in the caption with bounding boxes to indicate their positions. The format of the bounding box is $[[x_0, y_0, x_1, y_1]]$, where $(x_0, y_0)$ and $(x_1, y_1)$ represent the coordinates of upper-left and lower-right corners which are normalized to $[000, 999]$. If multiple objects are indicated by a single noun phrase, their boxes are separated by semicolons in double square brackets. 
% We have also collected grounding data on web page elements, which will be introduced in the next part. 

\noindent \textbf{GUI imagery.} %Datasets such as LAION and COYO are mainly composed of natural images, in which GUI images are fewer and often weakly relevant to the corresponding text. Concurrently, GUI images differ significantly from natural images and possess unique elements (e.g., input fields, hyperlinks, icons) and characteristics (such as the layout of interface elements). To enhance the model's understanding of GUI images during the pre-training phase, we innovatively design two GUI grounding tasks, i.e., 
Our approach innovatively addresses the scarcity and limited relevance of GUI images in datasets like LAION and COYO, which predominantly feature natural images. GUI images, with their distinct elements such as input fields, hyperlinks, icons, and unique layout characteristics, require specialized handling. To boost the model's capability in interpreting GUI imagery, we have conceptualized two pioneering GUI grounding tasks: (1) GUI Referring Expression Generation (REG) – where the model is tasked with generating HTML code for DOM (Document Object Model) elements based on a specified area in a screenshot, and (2) GUI Referring Expression Comprehension (REC) – which involves creating bounding boxes for given DOM elements.
To facilitate robust training in GUI grounding, we have constructed the CCS400K (Common Crawl Screenshot 400K) dataset. This extensive dataset is formed by extracting URLs from the latest Common Crawl data, followed by capturing 400,000 web page screenshots. Alongside these screenshots, we compile all visible DOM elements and their corresponding rendered boxes using Playwright\footnote{\url{https://playwright.dev}}, supplementing the dataset with 140 million REC and REG question-answer pairs. This rich dataset ensures comprehensive training and understanding of GUI elements. To mitigate the risk of overfitting, we employ a diverse range of screen resolutions for rendering, selected randomly from a list of commonly used resolutions across various devices. Additionally, to prevent the HTML code from becoming overly extensive and unwieldy, we perform necessary data cleaning by omitting redundant attributes in the DOM elements, following the method outlined in ~\cite{lee2023pix2struct}.
%However, designing self-supervised pre-training tasks for web pages is not straightforward. A direct approach involves pairing screenshots with HTML~\cite{lee2023pix2struct}, but even after extensive manual rule-based cleaning and filtering, the resulting HTML remains lengthy, which leads to unsustainable training overheads and hinders the model's ability to learn fine-grained associations between screenshots and HTML. 
 % with (1) grounded captioning, i.e. each entity in the caption is followed by corresponding bounding box, (2) Referring Expression Comprehension (REC) tasks (i.e. to predict the object's bounding box given the text description) and 

We also incorporate publicly available text-image datasets including LAION-2B and COYO-700M (after removing the broken URLs, NSFW images, and images with noisy captions and political bias) during pre-training. 

We pre-train our CogAgent model for a total of 60,000 iterations with a batch size of 4,608 and a learning rate of 2e-5. We freeze all parameters except the newly added high-resolution cross-module for the first 20,000 steps, resulting in a total number of 646M (3.5\%) trainable parameters, then additionally unfreeze the visual expert in CogVLM for the next 40,000 steps. We warm up with curriculum learning by first training on easier text recognition (synthetic renderings and OCR on natural images) and image captioning, then sequentially incorporating harder text recognition (academic document), grounding data and web page data, as we observed that it leads to faster convergence and more stable training in our preliminary experiments.  

% data amount
% and the pretraining consumes about 1,220 A800×days.

\subsection{Multi-task Fine-tuning and Alignment}
To enhance our model's performance for diverse tasks and ensure it aligns with free-form human instructions in the GUI setting, we further fine-tune our model on a broad range of tasks. We manually collected over two thousand screenshots from mobile phones and computers, each annotated with screen elements, potential tasks, and methods of operation in the question-answering format by human annotators (details illustrated in the Appendix). We also utilize Mind2Web~\cite{deng2023mind2web} and AITW~\cite{rawles2023android}, datasets focusing on web and Android behaviors which comprise tasks, sequences of actions and corresponding screenshots, and convert them into a natural language question-and-answer format using GPT-4. Besides, we incorporate multiple publicly available visual question-answering (VQA) datasets encompassing a variety of tasks into our alignment dataset. We unfreeze all model parameters during this stage and train for 10k iterations with a batch size of 1024 and a learning rate of 2e-5. 
% , including VQAv2, OK-VQA, TextVQA, OCRVQ, STVQA, ChartQA, DocQA, etc., 
% and format the training instructions as ``\texttt{Question: <question> \textbackslash n Short Answer: <answer>}''

% \subsection{Multi-task Fine-tuning and Alignment}
% To further fine-tune our model to diverse tasks and align with free-form human instructions, we sequentially conducted fine-tuning in the following two stages:

% \textbf{Multi-task fine-tuning}. We collect multiple publicly available visual question answering (VQA) datasets encompassing a variety of tasks, including VQAv2, OK-VQA, TextVQA, OCRVQ, STVQA, ChartQA, DocQA, etc., and format the training instructions as ``\texttt{Question: <question> \textbackslash n Short Answer: <answer>}''. We unfreeze all model parameters and train for \hwy{iter} iterations with a batch size of \hwy{bsz} and a learning rate of 1e-5. 

% \textbf{Alignment}. To enhance the alignment of the model with user instructions in the GUI scenario, we manually collected over two thousand screenshots from mobile phones and computers. These screenshots were annotated to identify screen elements, potential tasks, and methods of operation in the question-answering format by human annotators (details are illustrated in the appendix\hwy{appendix}). We also utilize Mind2Web~\cite{deng2023mind2web}, a web behavior dataset containing webpage screenshots, and transform the tasks and their corresponding past and future actions into natural language question-and-answer format using GPT-4. \hwy{hyper-params}
 \begin{table*}[ht]
  \vspace*{-0.1cm}
 % \caption{Performance comparison of state-of-the-art visual question answering methods on various datasets. * refers to few-shot or zero-shot setting. }

  \resizebox{\textwidth}{!}{%
  \centering
  \renewcommand{\arraystretch}{1.15}
  \setlength{\tabcolsep}{10pt}
  \ 
  {
    \begin{tabular}{lccccccccc}
    \toprule
    % \multirow{2}{*}{\textbf{Method}} & \multicolumn{2}{p{0.4cm}}{\textbf{VQAv2}} & \multicolumn{1}{r}{\textbf{OKVQA}} & \multicolumn{1}{c}{\textbf{GQA}} & \multicolumn{2}{l}{\textbf{VizWizQA}} & \multicolumn{1}{c}{\textbf{OCRVQA}} & \multicolumn{1}{c}{\textbf{TextVQA}} & \textbf{SciQA}\\
    % \cmidrule(r{10pt}){2-3} \cmidrule(r){4-4} \cmidrule(r){5-5} \cmidrule(r){6-7} \cmidrule(r){8-8} \cmidrule(r){9-9} \cmidrule(r){10-10}
    % & test-dev & test-std & val & test-balanced & test-dev & test-std & test & test & IMG \\
        \multirow{2}{*}{\textbf{Method}} & \multicolumn{2}{c}{\textbf{General VQA}} & \multicolumn{6}{c}{\textbf{Text-rich VQA}}\\
    \cmidrule(r){2-3} \cmidrule(r){4-9}
    & VQAv2 & OKVQA  & OCRVQA & TextVQA & STVQA & ChartQA & InfoVQA & DocVQA \\
    \addlinespace[-0.1mm]
        \midrule
        \addlinespace[-0.2mm]
        \multicolumn{3}{l}{\color{dtdark}\it{\textbf{task-specific fine-tuning models}}} \\
        \hline
        \color{dt}Pix2Struct~\cite{lee2023pix2struct} & \color{dt}- & \color{dt}- & \color{dt}- & \color{dt}- & \color{dt}- & \color{dt}58.6 & \color{dt}40.0 & \color{dt}76.6\\
        \color{dt}BLIP-2~\cite{li2023blip} & \color{dt}82.2 & \color{dt}59.3 & \color{dt}72.7 & \color{dt}- & \color{dt}- & \color{dt}- & \color{dt}- & \color{dt}-\\
        
        \color{dt}PALI-X-55B~\cite{chen2023pali} & \color{dt}\underline{86.0} & \color{dt}\underline{66.1} & \color{dt}75.0 & \color{dt}71.4 & \color{dt}79.9 & \color{dt}\underline{70.9} & \color{dt}\underline{49.2} & \color{dt}80.0\\
        \color{dt}CogVLM{\tiny task-specific}~\cite{wang2023cogvlm} & \color{dt}84.7 & \color{dt}64.7 & \color{dt}74.5 & \color{dt}69.7 & \color{dt}- & \color{dt}- & \color{dt}- & \color{dt}- \\
        \hline
        \multicolumn{3}{l}{\it{\textbf{generalist models}}} \\
        \hline
        UReader~\cite{ye2023ureader} & - & 57.6 & - & - & - & 59.3 & 42.2 & 65.4\\
        Qwen-VL~\cite{bai2023qwen} & 79.5 & 58.6 & \underline{\textbf{75.7}} & 63.8 & - & 65.7 & - & 65.1\\
        Qwen-VL-chat~\cite{bai2023qwen} & 78.2 & 56.6 & 70.5 & 61.5 & - & 66.3 & - & 62.6 \\
        Llava-1.5~\cite{liu2023improved} & 80.0 & - & - & 61.5 & - & - & - & - \\
        Fuyu-8B~\cite{fuyu-8b} & 74.2 & 60.6 & - & - & - & - & - & - \\
        CogVLM{\tiny generalist}~\cite{wang2023cogvlm} & 83.4 & 58.9 & 74.1 & 68.1 & - & - & - & - \\
        CogAgent (Ours) & \textbf{83.7} & \textbf{61.2} & 75.0  & \underline{\textbf{76.1}} & \underline{\textbf{80.5}} & \textbf{68.4} & \textbf{44.5} &  \underline{\textbf{81.6}}\\

    \bottomrule
    \end{tabular}
  }
 }
   \vspace{-0.2cm}

   \caption{\textbf{Performance on Visual Question Answering benchmarks.} Bold text indicates the best score among the generalist category, and underlined text represents the best score across both generalist and task-specific categories.}
    \vspace{-3mm}

  \label{table:vqa}
\end{table*}

\section{Experiments}

To evaluate the foundational capabilities and GUI-related performance of our model, we conduct extensive experiments on a broad range of datasets. First, we conduct evaluations on eight VQA benchmarks, as well as MM-Vet~\cite{yu2023mm} and POPE~\cite{li2023evaluating}, which validate our model's enhanced ability in visual understanding, especially on those that are reliant on text recognition. Then we evaluate our model on Mind2Web and AITW datasets, as the representative of two major GUI scenarios --- computers and smartphones. 
% \hwy{Where to put image demo?}

\subsection{Foundational Visual Understanding}

We first extensively evaluate CogAgent's foundational visual understanding capability across eight VQA benchmarks, covering a wide range of visual scenes. The benchmarks can be divided into two categories: general VQA, including VQAv2~\cite{antol2015vqa} and OK-VQA~\cite{marino2019ok}, and text-rich VQA, including TextVQA~\cite{singh2019towards}, OCR-VQA~\cite{mishra2019ocr}, ST-VQA~\cite{biten2019scene}, DocVQA~\cite{mathew2021docvqa}, InfoVQA~\cite{mathew2022infographicvqa} and ChartQA~\cite{masry2022chartqa}. The latter category emphasizes the understanding of visually-situated text, including documents, charts, photographs containing text, etc. 
% Detailed information for all datasets is provided in the Appendix.
% Contrary to models individually fine-tuned for optimal performance on each downstream task, our model is fine-tuned collectively on all datasets simultaneously, yielding a single generalist model which is then evaluated across all datasets. 
% The goal of generalist evaluation is to better mirror real-world situations of visual agents where typically a single model is used, and to demonstrate the model's versatility and robustness across tasks. 
To demonstrate the model's versatility and robustness across tasks, our model is fine-tuned collectively on all datasets simultaneously, yielding a single generalist model which is then evaluated across all datasets.

% report generalist + why
The results are presented in \cref{table:vqa}. For general VQA, CogAgent achieves state-of-the-art generalist results on both datasets. For text-rich VQA, CogAgent achieves state-of-the-art results on 5 out of 6 benchmarks, significantly surpassing generalist competitors (TextVQA+8.0, ChartQA+2.1, InfoVQA+2.3, DocVQA+16.2), even outperforming the task-specific state-of-the-art models on TextVQA(+4.7), STVQA(+0.6) and DocVQA(+1.6). Notably, compared to the generalist results of CogVLM which CogAgent is initially based on, CogAgent demonstrates certain improvements on both general and Text-rich VQA tasks, suggesting the efficacy of our proposed model architecture and training methods. 

\begin{table}[b]

    \centering
    \vspace{-0.3cm}
    \setlength{\tabcolsep}{10pt}
    \resizebox{\columnwidth}{!}{
    \begin{tabular}{@{}llcc@{}}
        \toprule
        \textbf{Method} & \textbf{LLM} & \textbf{MM-Vet}  & \textbf{POPE{\tiny adv}}\\
        \midrule
        BLIP-2~\cite{li2023blip}& Vicuna-13B & 22.4 & -\\	
        Otter~\cite{li2023mimic} & MPT-7B& 24.7 & -\\
        MiniGPT4~\cite{zhu2023minigpt} & Vicuna-13B & 24.4 & 70.4 \\
        InstructBLIP~\cite{dai2023instructblip} & Vicuna-13B & 25.6 & 77.3\\
        LLaVA~\cite{liu2023visual}& LLaMA2-7B & 28.1 & 66.3\\
        LLaMA-Adapter v2~\cite{gao2023llamaadp}& LLaMA-7B & 31.4 & -\\
         DreamLLM~\cite{dong2023dreamllm} & Vicuna-7B & 35.9 & 76.5\\
        LLaVA-1.5~\cite{liu2023improved}& Vicuna-13B & 36.3 & 84.5\\
        Emu~\cite{sun2023generative} &LLaMA-13B & 36.3 & -\\
        % Qwen-VL-chat~\cite{bai2023qwen} & Qwen-7B & - & 1487.6 \\
        CogAgent (Ours) & Vicuna-7B & \textbf{52.8} & \textbf{85.9}\\
        \bottomrule
    \end{tabular}
    }
    \vspace{-0.1cm}
    \caption{\textbf{Evaluation of CogAgent on conversational style QA and hallucination assessment.} Regarding the POPE dataset, we use its adversarial subset for this evaluation. }
    \label{tab:LLaVA_results}
    \vspace{-0.1cm}
\end{table}

Furthermore, we conducted zero-shot tests of our model on the challenging MM-Vet~\cite{yu2023mm} and POPE~\cite{li2023evaluating} datasets, both of which are instrumental in gauging the multi-modal capabilities and the generalization performance in complex tasks including conversation
question-answering, detailed descriptions, complex reasoning
tasks. MM-Vet is designed with six core tasks to assess multi-modal models' proficiency in handling intricate assignments, and POPE-adversarial models on their susceptibility to hallucinations. Our experimental results, as detailed in Table~\ref{tab:LLaVA_results}, showcase that our model significantly outperforms other existing models in both datasets. Notably, on the MM-Vet dataset, our model achieved a remarkable score of 52.8, surpassing the closest competitor, LLaVA-1.5, by a substantial margin (+16.5). On the POPE-adversarial evaluation, our model attained a score of 85.9, demonstrating superior handling of hallucinations compared to other models.

% These results indicate CogAgent's robust performance in foundational visual understanding, especially in the interpretation of images with embedded text. With these core competencies, the model can be feasibly applied to various visual agent tasks across different GUI environments.

\subsection{GUI Agent: Computer Interface}

\begin{table}[t]
\centering
% \resizebox{0.98\columnwidth}{!}{

\resizebox{\columnwidth}{!}{
\small{
\setlength{\tabcolsep}{2pt}

\begin{tabular}{@{}l|ccc|c@{}}
\toprule
\textbf{Method} & \textbf{cross-task} & \textbf{cross-website} & \textbf{cross-domain} & \textbf{overall} \\
\midrule
\addlinespace[-0.1mm]
    \multicolumn{3}{l}{\it{Representations of screen inputs: HTML}} \\
\addlinespace[-0.7mm]
\midrule
GPT-3.5\cite{openai2022chatgpt}{\tiny (few-shot)} & 18.6 & 17.4 & 16.2 & 17.4 \\
GPT-4\cite{openai2023gpt4}\textsuperscript{\dag}{\tiny (few-shot)} & 36.2 & 30.1 & 26.4 & 30.9 \\
Flan-T5{\tiny XL} \cite{deng2023mind2web} & 52.0 & 38.9 & 39.6 & 43.5 \\
LLaMA2-7B\cite{touvron2023llama} & 52.7 & 47.1 & 50.3 & 50.1 \\
LLaMA2-70B\cite{touvron2023llama} & 55.8 & 51.6 & 55.7 & 54.4 \\
% MindAct{\footnotesize(Flan-T5{\tiny XL})}
% LLaMA2-70B & --- & --- & --- & 13.5  \\

% GPT 4 in Agenttuning & & & & 22.6 \\ 
% GPT 3.5 in Agenttuning & & & & 15.7 \\
\midrule
\addlinespace[-0.1mm]
    \multicolumn{3}{l}{\it{Representations of screen inputs: Image}} \\
\addlinespace[-0.7mm]
\midrule
Qwen-VL\cite{bai2023qwen} & 12.6 & 10.1 & 8.0 & 10.2 \\
CogVLM\cite{wang2023cogvlm} & 37.1 & 23.4 & 26.3 & 23.9 \\
CogAgent (Ours) & \textbf{62.3} & \textbf{54.0} & \textbf{59.4} & \textbf{58.2} \\

\bottomrule
\end{tabular}
}
}
\label{tab:mind2web}
\vspace{-0.15cm}

\caption{\textbf{Performance on Mind2Web.} \dag ~denotes element selection from top-10 element candidates, others from top-50, following \citet{deng2023mind2web}. Results for GPT-3.5 and GPT-4 are from \citet{deng2023mind2web}.}
\vspace{-0.5cm}

\end{table}

We evaluate CogAgent on Mind2Web, a dataset for web agents that includes over 2,000 open-ended tasks collected from 137 real-world websites across 31 domains. 
% Each entry in the dataset comprises a high-level task description, a sequence of actions, and webpage snapshots in a variety of formats, including HTML and screenshots. 
Given the task description, current webpage snapshot and previous actions as inputs, agents are expected to predict the subsequent action. We follow the setting of \citet{deng2023mind2web} in our experiments, and report step success rate (step SR) metric. 
% Further details are attached in the Appendix.
% and formalize the problem as a two-step process: first choose the target webpage element among top-k (k=10 or 50) candidates, then predict specific operations. Further details are attached in the Appendix. It is noteworthy that our model possesses the capability to directly predict target elements and operations, which is demonstrated in \cref{fig:demo} and Appendix. 

% \hwy{Appendix: mind2web data; problem formulation}

Several language models were evaluated on this benchmark. For instance, AgentTuning~\cite{zeng2023agenttuning} and MindAct~\cite{deng2023mind2web} evaluated Llama2-70B and Flan-T5-XL in a fine-tuned setting, and GPT-3.5 and GPT-4 in a in-context learning setting. However, limited by the input modality of language models, these models could only use heavily cleansed HTML as the representation of screen inputs. To the best of our knowledge, no visually-based web agents have been experimented with on this benchmark. 
% \hwy{update baseline}

We fine-tune our model on the train set and evaluate on three out-of-domain subsets, i.e. cross-website, cross-domain, and cross-task. We additionally fine-tune LLaMA2-7B and LLaMA2-70B as the baseline of fine-tuned LLMs, and adopt the same HTML cleansing process as \citet{deng2023mind2web} to construct HTML input. The results are presented in \cref{tab:mind2web}. Compared to other methods, our approach achieved significant performance improvements across all three subsets, surpassing LLaMA2-70B, which is nearly 4$\times$ the scale of CogAgent, by 11.6\%, 4.7\%, and 6.6\%, respectively. This reflects not only the capability of our model but also the advantages of employing a visual agent in computer GUI scenarios. 
% \hwy{zero-shot of our model?}

% Samples in \hwy{todo: demo} further demonstrate our model capability in Computer GUI.

% We use the same evaluation pipeline as \citet{deng2023mind2web} and \citet{zeng2023agenttuning} and test on 3 out-of-domain test sets (cross-website, cross-domain, cross-task), and report step success rate (step SR) as metric. 

% Following \citet{deng2023mind2web} and \citet{zeng2023agenttuning}, we report step success rate (step SR) as metric, and formalize the problem as: first choose the target webpage element among top-k (k=10 or 50) candidates, then predict specific operations. It is noteworthy that our model possesses the capability to directly predict target elements and operations, which is demonstrated in \hwy{demo img} and Appendix. 

% Unlike datasets relying on simulated or overly simplified web data, Mind2Web utilizes real-world websites and is annotated by human annotators. 

\subsection{GUI Agent: Smartphone Interface}
\begin{table}[t]
\centering
\setlength{\tabcolsep}{2pt}

\resizebox{\columnwidth}{!}{
\small{
\begin{tabular}{@{}l|ccccc|c@{}}
\toprule
\textbf{Method} & \textbf{GoogleApp} & \textbf{Install} & \textbf{WebShop} & \textbf{General} & \textbf{Single} & \textbf{Overall} \\ 

\midrule
\addlinespace[-0.1mm]
    \multicolumn{7}{l}{\it{Representations of screen inputs: textual description (OCR+icon)}} \\
\addlinespace[-0.7mm]
\midrule

GPT-3.5\cite{openai2022chatgpt}{\footnotesize{(few-shot)}} & 10.47 & 4.38 & 8.42 & 5.93 & 9.39 & 7.72 \\ 
LLaMA2-7B\cite{touvron2023llama}\textsuperscript{\dag}  & 30.99 & 35.18 & 19.92 & 28.56 & 27.35 & 28.40 \\ 

\midrule
\addlinespace[-0.1mm]
    \multicolumn{4}{l}{\it{Representations of screen inputs: image}} \\
\addlinespace[-0.7mm]
\midrule

% BC-history\cite{rawles2023android}\textsuperscript{\dag} & 75.7 & 77.5 & 68.5 & 63.7 & 80.3 & 73.1 \\
Auto-UI{\tiny unified}\cite{zhan2023you}  & 71.37 & 76.89 & 70.26 & \textbf{68.24} & 84.58 & 74.27 \\ 
CogAgent (Ours) & \textbf{74.95} & \textbf{78.86} & \textbf{71.73} & 65.38 & \textbf{93.49} & \textbf{76.88} \\ 
\bottomrule
\end{tabular}
}
}
    \vspace{-0.1cm}

\caption{\textbf{Performance on Android in the Wild (AITW) dataset.} \dag~ represents models individually fine-tuned on each subset, while others are unified models across all subsets. The results of LLaMA2 and GPT-3.5 are from \citet{zhan2023you}.}

\label{tab:aitw}
\vspace{-0.2cm}
\end{table}

To evaluate our model on diverse smartphone interfaces and tasks, we utilize Android in the Wild (AITW) dataset~\cite{rawles2023android} , a large-scale dataset for Android device agents. 
% It comprises 715k operation episodes, covering 30k distinct task instructions, four Android versions, and eight device types featuring varying screen resolutions. 
It comprises 715k operation episodes covering varying Android versions and device types.
Each episode in the dataset consists of a goal described in natural language, followed by a sequence of actions and corresponding screenshots. The training target is to predict the next action based on the given goal, historical actions, and the screenshot.
% AITW considers a wide range of action types, including tapping, swiping, typing, going home, going back, entering, etc. 
For each action, models are required to predict the exact action type; for tap, swipe and type, models are further required to predict the position, direction, and content to be typed, respectively. 

We conduct comparisons with two kinds of baselines: language models using the textual description of UI elements provided by the original dataset (text OCR and icon) as the representations of screen inputs\footnote{Some Android applications may have View Hierarchy which is more friendly to language-based agents, but most of them tend to be poor quality or missing altogether. Therefore, as a large-scale, general-purpose dataset, AITW retained the results of OCR detection and icon detection as textual representations of screenshots.}, and visual-language models using images as the screen inputs. 
% LLaMA2 is specifically fine-tuned on each subset, while GPT-3.5 is assessed using 5-shot in-context learning. 
We simultaneously fine-tuned on all the subsets, yielding a unified model which is then evaluated on all test sets. As the GoogleApps subset is 10-100 times larger than other subsets, we downsample it to 10\% to avoid data imbalance. 

Results are shown in \cref{tab:aitw}. CogAgent achieves state-of-the-art performance compared to all previous methods. In comparison to language-based methods, our model surpasses both baselines by a large margin. In comparison to the visual-language baseline, Auto-UI, our model achieves +2.61 improvements in the overall performance. In instances of inaccuracies, we randomly sample hundreds of cases, and upon reassessment, more than 40\% are determined to be correct (refer to the appendix for details). This diversity arises from the multiple valid pathways inherent in mobile interactions, resulting in a range of responses.
% \hwy{more info:  The textual description of the screen is formatted as an HTML syntax (write in Appendix?)}

 \section{Ablation Study} \label{subsec:ablation}

To thoroughly comprehend the impact of various components in the methodology, we conduct ablation studies on two aspects, model architecture and training data. The evaluation is conducted on diverse datasets, including multiple VQA datasets (STVQA, OCRVQA, DocVQA) 
and a web agent dataset (Mind2Web).
For VQA datasets, we fine-tune the model on four datasets together for 3,000 iters with a batch size of 1,280, and report the generalist score; for Mind2Web, models are fine-tuned for 2,400 iters with a batch size of 128 and use top-10 setting. Training iterations are fewer than those in the main experiment, aiming to control variables within the constraints of a limited budget.

\subsection{Model Architecture}

To ascertain the efficacy of the high-resolution cross-module, we compare it with directly increasing the resolution using the original model architecture of CogVLM, and ablate on two perspectives: computational efficiency and model performance. 

\begin{figure}[h]
    \centering
    \vspace{-0.15cm}
    \includegraphics[width=0.9\linewidth]{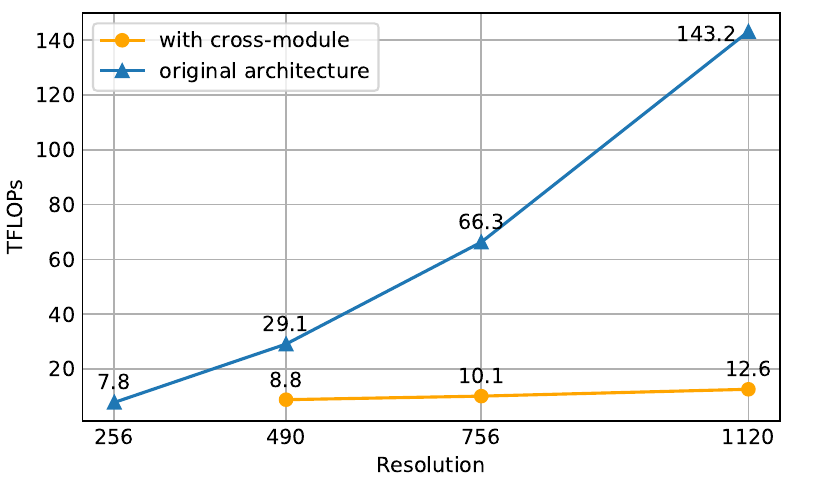}
        \vspace{-0.1cm}

    \caption{Comparison of FLOPs during forward propagation for different model architectures and resolutions.}
    \vspace{-0.2cm}
    \label{fig:flop}
\end{figure}

To measure computational overhead, we use floating point operations (FLOPs) as the metric, and conduct experiments on multiple resolutions including 224, 490, 756, and 1120. From \cref{fig:flop} we can see that, as the image resolution increases, models that use a high-resolution cross-module experience only a modest rise in computational overhead, demonstrating an almost linear relationship with the number of image patches. In contrast, using the original model structure, i.e. CogVLM, leads to a significant increase in the number of FLOPs at higher resolutions. Its FLOPs can even be more than 10 times higher compared to employing a cross-module at a resolution of 1120, which is the resolution utilized by CogAgent.

We further compare the model performance in \cref{tab:ablation-architecture}, which indicates that models with high-resolution cross-module at the resolution of 756 require only 1/2 of the computational resources used by the original structure at the resolution of 490, while delivering significantly better performance. Additionally, the high-resolution cross-module allows for further increasing models' acceptable resolution within a limited computational budget, thereby yielding additional performance improvements.

\begin{table}[h]
% \vspace*{-0.1cm}

% \xjz{To complete the ablation experiment, the control variables were set to the same number of training steps, and thus a different number of steps than the main experiment. This is a more convincing explanation than a smaller number of steps.}

\setlength{\tabcolsep}{1pt}
\resizebox{\columnwidth}{!}{
\begin{small}
\begin{tabular}{c|cc|cccc|cc}
\toprule
high-res  & base   & cross  & \multirow{2}{*}{STVQA} & \multirow{2}{*}{OCRVQA} & \multirow{2}{*}{DocVQA} & \multirow{2}{*}{Mind2Web}  & training & \multirow{2}{*}{TFLOPs}\\
 module & res & res &  & & & & time/it (s) &  \\
\midrule

\XSolidBrush & 224 & ---  & 48.0	& 70.2	& 28.6	& 34.6 & 2.36 & 7.77\\
\XSolidBrush & 490 & ---  & 68.1	& 74.5  	& 57.6	& 40.7 & 6.43 & 29.14\\
% \XSolidBrush & 756 & --- & 79.5	& 74.5	& 76.1 & 44.1 &  62&\\
 
\Checkmark & 224 & 756 & 73.6 & 74.2 & 62.3 & 40.7 &  3.57 & 10.08\\
\Checkmark & 224 & 1120 & 78.2 & 75.9 & 74.1 & 41.4 & 5.17 & 12.56\\

\bottomrule
\end{tabular}%
\end{small}}
% }
\vspace*{-0.1cm}
\caption{{Ablation study on model architecture. Training time is evaluated on A800 with the batch size of 8. Models are pre-trained with Caption+OCR data. } 
}
\vspace*{-0.4cm}
\label{tab:ablation-architecture}
\end{table}

\subsection{Pre-train Data}

\begin{table}[h]

\setlength{\tabcolsep}{1pt}

\resizebox{\columnwidth}{!}{

\begin{small}
\begin{tabular}{c|cc|ccccc}
\toprule
pre-train data  & base res   & cross res  & STVQA & OCRVQA & DocVQA  & Mind2Web \\
% pre-train  & base   & cross  & STVQA & OCRVQA & DocVQA  & Mind2Web \\
% data & res & res &  & & & & \\
\midrule

% Cap & 756  & ---  & 75.3  & 75.1	& 72.4 & ---	\\
% Cap+OCR & 756 & --- & 79.5	& 74.5	& 76.1  & 44.0 \\

Cap & 490  & ---  & 68.1	& 74.5	& 57.6 & 38.6	\\
Cap+OCR & 490 & --- & 72.5	& 75.0	& 59.8  & 40.7 \\

\midrule 

Cap+OCR & 224 & 1120 & 78.2 & 75.9 & 74.1  & 41.4 \\
All & 224 & 1120 & 79.4 & 75.6 & 76.4  & 54.2 \\

\bottomrule
\end{tabular}%
\end{small}
}
\vspace*{-0.15cm}

\caption{Ablation study on pre-train data with sequentially added image captioning, OCR and other pre-train data. 
% \xjz{May need explanation why there's no value the same as which in main table.}
}
\vspace*{-0.2cm}
\label{tab:ablation-data}
\end{table}

%\hwy{caution footnote label!!!}
\vspace{-0.2cm}
We further conduct an ablation study on pre-training data, which is an integral part of training visual agents. Building upon the image-caption data commonly used in visual-language training, we sequentially add OCR data (denoted as Cap+OCR), as well as GUI and grounding data (denoted as All). The results in \cref{tab:ablation-data} indicate that each part of data broadly contributes to enhanced performance. Notably, web and grounding data have a significant impact on the Mind2Web dataset, underscoring the importance of constructing domain-specific pre-train data in the training of GUI agents.

 \section{Conclusion}
We introduce CogAgent, a VLM-based GUI agent with enhanced pre-train data construction and efficient architecture for high-resolution input. CogAgent achieves state-of-the-art performance on a wide range of VQA and GUI benchmarks, and will be open-sourced.  
CogAgent is an initial exploration of VLM-based GUI agent, and still has some shortcomings, e.g. imprecise output coordinates and incapability of processing multiple images, necessitating further research.

 % \clearpage
 \subsubsection*{Acknowledgments}
This work is supported by Technology and Innovation Major Project of the Ministry of Science and Technology of China under Grant 2022ZD0118600, Natural Science Foundation of China (NSFC) 62277033 and the New Cornerstone Science Foundation through the XPLORER PRIZE. It also got partial support from the National Engineering Laboratory for Cyberlearning and Intelligent Technology, Beijing Key Lab of Networked Multimedia, Daimler Greater China Ltd. -Tsinghua University Joint Institute for Sustainable Mobility, Tsinghua University(Department of Computer Science and Technology)-Siemens Ltd., China Joint Research Center for Industrial Intelligence and Internet of Things (JCIIOT) and a research fund from Zhipu AI.
 
% We thank Xiaohan Zhang from Zhipu AI for managing the data annotation team, Bin Xu, Juanzi Li from Tsinghua for guidance, and Zhao Xue, Aohan Zeng, Yifan An, Chenxu Guo from Zhipu AI and Tsinghua for data management.

 {
     \small
     \bibliographystyle{ieeenat_fullname}
     \bibliography{main}
 }

% WARNING: do not forget to delete the supplementary pages from your submission 
\clearpage
\setcounter{section}{0}
% \maketitlesupplementary

\twocolumn[
        \centering
        \Large
        \textbf{\thetitle}\\
        \vspace{0.5em}Appendix \\
        \vspace{1.0em}
       ]
\section{Details of Training Configurations}
We report the detailed training settings of CogAgent in \Cref{tab:appendix_hyperparam_pretrain}, and model configurations of CogAgent in \Cref{tab:appendix_model_config}.

\begin{small}
\begin{table}[htbp]
\centering
    \renewcommand{\arraystretch}{1.15}

    \setlength{\tabcolsep}{0.35mm}{
    \resizebox{0.9\columnwidth}{!}{
    \setlength{\tabcolsep}{10pt}
    \begin{tabular}{ccc}
    \toprule
    \textbf{Configurations} & \textbf{Pre-train} & \textbf{Multi-task} \\
      \midrule
      Total steps & $60,000$ & $10,000$\\
      Warmup steps & $500$ & $500$  \\
      Batch size & $4,608$ & $1,024$\\
      Learning rate & \multicolumn{2}{c}{$2\times 10^{-5}$}  \\
      Learning rate decay & \multicolumn{2}{c}{$\operatorname{Cosine}$} \\
      Weight decay & \multicolumn{2}{c}{0.05} \\
      Dropout ratio & \multicolumn{2}{c}{$0.1$} \\
      Adam $\epsilon$ & \multicolumn{2}{c}{$1\times 10^{-5}$}  \\
      Adam $\beta$ & \multicolumn{2}{c}{(0.9, 0.95)} \\
    \bottomrule
    \end{tabular}}
    }
    \caption{
        Training settings of pre-training and multi-task fine-tuning. 
    }
    \vspace{2mm}
  \label{tab:appendix_hyperparam_pretrain}
\end{table}
\end{small}

\begin{small}
\begin{table}[htbp]
\centering
    \renewcommand{\arraystretch}{1.15}

    \setlength{\tabcolsep}{0.35mm}{
    \resizebox{0.9\columnwidth}{!}{
    \setlength{\tabcolsep}{7pt}
    \begin{tabular}{ccc}
    \toprule
      \multicolumn{3}{l}{\textit{\textbf{VLM decoder}}}\\
      \hline
      Architecture &  \multicolumn{2}{c}{Vicuna-1.5-7B + visual expert}\\
      % Parameters &  \multicolumn{2}{c}{12B}\\
      Layers & \multicolumn{2}{c}{32} \\
      Hidden size & \multicolumn{2}{c}{$4,096$} \\
      Attention heads & \multicolumn{2}{c}{$32$} \\
      \hline
      \multicolumn{3}{l}{\textit{\textbf{Low-resolution visual encoder}}}\\
      \hline
      
      Architecture &  \multicolumn{2}{c}{EVA2-CLIP-E}\\
      % Parameters &  \multicolumn{2}{c}{5B}\\
      Input resolution & \multicolumn{2}{c}{$224\times224$}  \\
      Patch size & \multicolumn{2}{c}{$14\times14$}  \\
      \hline
      \multicolumn{3}{l}{\textit{\textbf{High-resolution visual encoder}}}\\
      \hline
      Visual encoder &  \multicolumn{2}{c}{EVA2-CLIP-L}\\
      % Parameters &  \multicolumn{2}{c}{0.43B}\\
      Input resolution & \multicolumn{2}{c}{$1120\times1120$}  \\
      Patch size & \multicolumn{2}{c}{$14\times14$}  \\

    \hline
      \multicolumn{3}{l}{\textit{\textbf{Cross Attention}}}\\
        \hline
      Hidden size & \multicolumn{2}{c}{$1,024$} \\
      Attention heads & \multicolumn{2}{c}{$32$} \\
      % \hline
    % Number of layers & $6$ \\
    %   Hidden size & $768$  \\
    %   FFN inner hidden size & $3072$  \\
    %   Number of attention heads & $12$  \\
    \bottomrule
    \end{tabular}}
    }
    \caption{
        Model configurations of CogAgent. 
    }
    \vspace{2mm}
  \label{tab:appendix_model_config}
\end{table}
\end{small}

\section{Details of Evaluation Datasets}
In this section, we will provide a detailed overview of the datasets used in our evaluations
% , along with explanations of the evaluation procedures
.

% Summary of Evaluation benchmarks

\begin{table*}[t]

\centering
% \resizebox{\textwidth}{!}{%
  \resizebox{0.98\textwidth}{!}{%
    \begin{tabular}{l|llll}
\toprule
\textbf{Task}                      & \textbf{Dataset} & \textbf{Description}                                & \textbf{Split}    & \textbf{Metrics}       \\ \midrule
\multirow{4}{*}{General VQA}       & VQAv2            & VQA on natural images.                              & test-dev          & VQA Score($\uparrow$)  \\
                                   & OK-VQA           & VQA on natural images requiring outside knowledge.  & val               & VQA Score ($\uparrow$) \\
                                   & MM-Vet            & Conversational style VQA on integrated capabilities.                              & test          & GPT-4 score($\uparrow$)  \\
                                   & POPE            & VQA for hallucination assessment. The adversarial setting is used.                              & test          & F1 score($\uparrow$)  \\\midrule
\multirow{6}{*}{Text-rich VQA} & OCR-VQA          & VQA on images of book covers.                       & test              & EM ($\uparrow$)        \\
                                   & TextVQA          & VQA on natural images containing text.              & test               & VQA Score ($\uparrow$) \\
                                   & ST-VQA          & VQA on natural images requiring textual understanding.              & test               & ANLS ($\uparrow$) \\
                                   & ChartQA          & VQA about charts with visual and logical reasoning.              & test               & VQA Score ($\uparrow$) \\
                                   & InfoVQA          & VQA on infographics.              & test               & ANLS ($\uparrow$) \\
                                   & DocVQA          & VQA on document images.              & test               & ANLS ($\uparrow$) \\ \midrule
\multirow{2}{*}{GUI Agent}         & Mind2Web          & Web behavior prediction given snapshots and historical actions.                  & test            & step SR ($\uparrow$)  \\
                                   & AITW         & Android behavior prediction given snapshots and historical actions.                  & test            & Matching Score ($\uparrow$)  \\ \midrule
\end{tabular}

}
\caption{Summary of the evaluation benchmarks.}

\label{tab:appendix_eval_benchmarks}

\end{table*}

\subsection{General VQA}
\begin{itemize}
    \item \textbf{VQAv2~\cite{antol2015vqa}.} VQAv2 is designed for visual question answering with natural images, covering a wide range of question types including yes/no, numerical counting, and more open-ended inquiries. The dataset comprised of a collection exceeding 200,000 images, paired with more than 1.1 million questions. Each question is paired with 10 answers annotated by different annotators. 
    \item \textbf{OK-VQA~\cite{marino2019ok}.} OK-VQA (Outside Knowledge Visual Question Answering) dataset is constructed to evaluate visual question-answering skills that require external knowledge. Models need to combine image content and common sense to answer questions. The dataset includes 14,055 open-ended questions, each accompanied by 5 ground truth answers.
    \item \textbf{MM-Vet~\cite{yu2023mm}.} MM-Vet is designed to evaluate the overall capability of generalist visual language models in a zero-shot manner. It integrates 6 core VL capabilities to solve complex tasks (including recognition, OCR, knowledge, language generation, spatial awareness, and math), and explores 16 distinct integrations resulting from the combination of these capabilities. As for evaluation metrics, it utilizes a language model-based evaluator tailored for open-ended responses.
    \item \textbf{POPE~\cite{li2023evaluating}.} POPE (Polling-based Object Probing Evaluation) is a dataset constructed to assess the object hallucination problem in large visual language models. It employs a polling-based object probing method, transforming hallucination assessment into a binary classification challenge. This is achieved by prompting large vision-language models (LVLMs) with straightforward Yes-or-No queries regarding the objects in question (for example, "Is there a car in the image?"). Our evaluation is conducted under the dataset's most challenging setting: the adversarial setting.
\end{itemize}

\subsection{Text-rich VQA}
\begin{itemize}
    \item \textbf{OCR-VQA~\cite{mishra2019ocr}.} OCR-VQA dataset comprises 207,572 images of book covers, paired with more than 1 million question-answer pairs. The questions inquire about book information including title, edition, year, author, and genre of the book, which requires text recognition and comprehension abilities. 

    \item \textbf{TextVQA~\cite{singh2019towards}.} TextVQA is a benchmark of visual reasoning based on text in images. Models need to incorporate the textual information in the images and reason over it to answer TextVQA questions. It comprises a total of 28,408 images and 45,336 questions. 

    \item \textbf{ST-VQA~\cite{biten2019scene}.} ST-VQA is designed to emphasize the significance of exploiting the semantic information present within images in textual form during the VQA process. It comprises tasks of diverse difficulties, for which recognizing the scene text and performing necessary reasoning is required to generate the answer. The dataset comprises 23,038 images sourced from multiple public datasets and 31,791 question-answer pairs. 

    \item \textbf{ChartQA~\cite{masry2022chartqa}.} ChartQA is a benchmark of question-answering about logical and visual reasoning on charts. It consists of 20,882 charts curated from four different online sources, 9,608 manual written questions, as well as 23,111 questions automatically generated with T5 according to human-written chart summaries. 

    \item \textbf{InfographicVQA(InfoVQA)~\cite{mathew2022infographicvqa}.} The task of InfoVQA is to answer questions centering on a given infographic image. The answers to most questions can be extracted from the given documents, while the answers to a small percentage of questions are not extractive. There are 5K Images collected from the Internet and 30K manually annotated questions in the dataset. 

    \item \textbf{DocVQA~\cite{mathew2021docvqa}.} DocVQA focuses on question-answering given a document image. The answer for questions is often a span of text from the given documents. There are 12K images and 50K manually annotated questions in the datasets. 

\end{itemize}

\subsection{GUI Agent}
CogAgent is evaluated on two GUI agent datasets, Mind2Web and Android in the Wild (AITW), corresponding to computer agent and smartphone agent respectively.

\begin{itemize}
\item \textbf{Mind2Web~\cite{deng2023mind2web}.}
Mind2Web is designed to develop and evaluate web agents capable of executing intricate tasks on various websites based on language directions. While existing datasets for web agents commonly rely on simulated or overly simplified web data, Mind2Web utilizes real-world websites and is annotated by human annotators. It gathers data from 137 websites covering 31 domains, and collects over 2,000 open-ended tasks, each accompanied by a crowdsourced action sequence. 

In mind2web, evaluated agents are asked to accomplish a designated task on a chosen website by performing a sequence of actions. Each instance (i.e. a specific task) in Mind2Web contains a task description, action sequence, and webpage snapshots. Each action in the sequence is a (Target element, Operation) pair, and Operation includes Click, Type (with additional value), and Select (with additional value). Each action is paired with the concurrent webpage snapshots in a variety of formats including raw HTML code, DOM tree, screenshot, etc. As for CogAgent, we choose screenshot images as the input representation of websites; as for other language-based agents, HTML is chosen as the input representation. 

Following \citet{deng2023mind2web} and \citet{zeng2023agenttuning}, we formalize the problem as: first choose the target webpage element among top-k (k=10 or 50) candidates, then predict specific operations. The top-k candidates are provided by the candidate generation model in \citet{deng2023mind2web}. Step success rate (step SR) is reported on 3 out-of-domain test sets (cross-website, cross-domain, cross-task) as metric. Only predictions with the same target element and operation as the ground truth are regarded as correct. 

\item \textbf{Android in the Wild (AITW)~\cite{rawles2023android}.}
AITW is constructed to develop and evaluate Android device-control systems that are capable of understanding and acting upon human natural language instructions by directly manipulating the device's user interface. This dataset significantly surpasses others in its category in terms of size, encompassing 715k episodes across 30k distinct instructions, and covering four Android versions (v10–13). It also includes eight types of devices, ranging from Pixel 2 XL to Pixel 6, each with different screen resolutions. AITW consists of five subsets: GoogleApps, Install, WebShopping, General, and Single. The subsets have distinct tasks, while are in the same data format. Each episode (i.e. action sequence) comprises three components: a goal instruction provided in natural language, a user action sequence, and a corresponding screenshots sequence. 

As for screenshots, AITW only provides screenshot images and does not provide tree-based representations of UI. This is because a large portion of them in Smartphone applications are of low quality or even do not exist, and adopting the tree-based representation would strongly limit agents' applications. For visual agents, screenshots are provided to the agents in image format; for language-model-based agents evaluated by \citet{zhan2023you}, the textual representations of OCR and icons formatted in HTML syntax are provided. As for actions, AITW considers a variety of action types including tapping, swiping, typing, going home, going back, entering, etc. For each action, models are required to predict the exact action type; for tap, swipe and type, models are further required to predict the position, direction, and content to be typed, respectively. The detailed standard for computing matching scores is provided in \citet{rawles2023android}.

\end{itemize}

\section{Derivation of Acceleration for High-Resolution Cross-Module}

Suppose that $L_{I_{\text{lo}}}$, $L_{I_{\text{hi}}}$ and $L_T$ are the lengths of the low-resolution image, high-resolution image and text sequences. Let $H_{\text{cross}}, H_{\text{dec}}$ be the number of attention heads in cross-attention and self-attention, and $d_{\text{cross}}, d_{\text{dec}}$ be the dimension of each attention head.

If using our high-resolution cross-module, the computational complexity of attention is
\begin{equation}
\begin{split}
\text{T}_{\text{improved}} = \mathbf{O}\bigl( &(L_{I_{\text{lo}}} + L_T) L_{I_{\text{hi}}} H_{\text{cross}} d_{\text{cross}} \\
&+ (L_{I_{\text{lo}}} + L_T)^2 H_{\text{dec}} d_{\text{dec}} \bigr).
\end{split}
\end{equation}

If not utilizing the high-resolution cross-module and directly substituting low-resolution images with high-resolution ones, the computational complexity would be
\begin{align}
\text{T}_{\text{original}} = \mathbf{O}\bigl((L_{I_{\text{hi}}} + L_T)^2 H_{\text{dec}} d_{\text{dec}} \bigr).
\end{align}

The reduction factor of the computational complexity in attention, $\text{T}_{\text{original}}/{\text{T}_{\text{improved}}} $, equals to
\begin{align}
% \small{
    % & \frac{\text{T}_{\text{original}}}{\text{T}_{\text{improved}}} \\
    % = 
    & \frac{(L_{I_{\text{hi}}} + L_T)^2 H_{\text{dec}} d_{\text{dec}}}{(L_{I_{\text{lo}}} + L_T) L_{I_{\text{hi}}} H_{\text{cross}} d_{\text{cross}}+(L_{I_{\text{lo}}} + L_T)^2 H_{\text{dec}} d_{\text{dec}}} \\   
    &=  \frac{L_{I_{\text{hi}}} + L_T}{L_{I_{\text{lo}}} + L_T} \frac{(L_{I_{\text{hi}}} + L_T) H_{\text{dec}} d_{\text{dec}}}{L_{I_{\text{hi}}} H_{\text{cross}} d_{\text{cross}}+(L_{I_{\text{lo}}} + L_T) H_{\text{dec}} d_{\text{dec}}} \\ 
    &=  \frac{L_{I_{\text{hi}}} + L_T}{L_{I_{\text{lo}}} + L_T} \frac{(L_{I_{\text{hi}}} + L_T) \frac{H_{\text{dec}} d_{\text{dec}}}{H_{\text{cross}} d_{\text{cross}}} }{L_{I_{\text{hi}}}+(L_{I_{\text{lo}}} + L_T) \frac{H_{\text{dec}} d_{\text{dec}}}{H_{\text{cross}} d_{\text{cross}}}} \label{eq:appendix1}
    % \\  
    % = & \frac{L_{I_{\text{hi}}} + L_T}{L_{I_{\text{lo}}} + L_T} \frac{\frac{H_{\text{dec}} d_{\text{dec}}}{H_{\text{cross}} d_{\text{cross}}} }{ \frac{L_{I_{\text{hi}}}}{(L_{I_{\text{hi}}} + L_T)} + \frac{(L_{I_{\text{lo}}} + L_T)}{(L_{I_{\text{hi}}} + L_T)}  \frac{H_{\text{dec}} d_{\text{dec}}}{H_{\text{cross}} d_{\text{cross}}}} 
    % }
\end{align}

\newcommand{\Lh}{L_{I_{\text{hi}}}}
\newcommand{\Ll}{L_{I_{\text{lo}}}}
\newcommand{\Lt}{L_T}

\begin{itemize}
\item \textbf{Case 1:} $L_{I_{\text{lo}}},L_T \ll L_{I_{\text{hi}}}$. 

Given that $L_{I_{\text{lo}}}$ is much smaller than $L_{I_{\text{hi}}}$, when $L_T$ also satisfies being much smaller than $L_{I_{\text{hi}}}$, both $\frac{L_{I_{\text{lo}}}}{L_{I_{\text{hi}}}}$ and $\frac{L_T}{L_{I_{\text{hi}}}}$ become first-order small quantities. If conducting a 0-th order approximation for the complexity reduction factor, we obtain:
\begin{align}
\frac{\text{T}_{\text{original}}}{\text{T}_{\text{improved}}} = & \frac{\Lh (1+\frac{L_T}{\Lh})}{\Ll + L_T} \frac{\Lh(1+ \frac{L_T}{\Lh}) \frac{H_{\text{dec}} d_{\text{dec}}}{H_{\text{cross}} d_{\text{cross}}} }
{\Lh(1+\frac{\Ll + L_T}{\Lh} \frac{H_{\text{dec}} d_{\text{dec}}}{H_{\text{cross}} d_{\text{cross}}})} \\
= & \frac{\Lh (1+\frac{L_T}{\Lh})}{\Ll + L_T} \frac{(1+ \frac{L_T}{\Lh}) \frac{H_{\text{dec}} d_{\text{dec}}}{H_{\text{cross}} d_{\text{cross}}} }
{1+\frac{\Ll + L_T}{\Lh} \frac{H_{\text{dec}} d_{\text{dec}}}{H_{\text{cross}} d_{\text{cross}}}} \\
\approx & \frac{\Lh}{\Ll+\Lt}
\end{align}

\item \textbf{Case 2:} Our implementation.

In our implementation, $d_{\text{cross}}=32$, $H_{\text{cross}}=32$, and we inherits $d_{\text{dec}}=128$, $H_{\text{dec}}=32$ from CogVLM-17B. Both high- and low-resolution encoders patchify images with $14\times14$-pixel patches, thus $L_{I_{\text{hi}}}=6400$, $L_{I_{\text{lo}}}=256$. 
Subsitute the values to \cref{eq:appendix1}, we obtain 
\begin{align}
\frac{\text{T}_{\text{original}}}{\text{T}_{\text{improved}}} = & \frac{6400+\Lt}{256+\Lt} \frac{4(6400+\Lt)}{6400+4(256+\Lt)}\\
> & \frac{6400+\Lt}{256+\Lt}
\end{align}

That is to say, our method leads to at least $\frac{L_{I_{\text{hi}}}+L_{T}}{L_{I_{\text{lo}}}+L_{T}} = \frac{6400+L_{T}}{256+L_{T}} \times$ acceleration which is a stringent lower bound. Taking pre-training as an example, with $L_T$ typically below $512$ in earlier stages, our high-resolution cross-module can get an over $25 \times$ reduction in computational cost.
\end{itemize}

\section{Performance Analysis on AITW}

Android in the Wild (AITW) dataset is a large-scale dataset for Android device agents. Our analysis centers on instances where the CogAgent's responses diverge from the annotated ground truth, focusing on these discrepancies.

\begin{figure}[h]
    \centering
    \includegraphics[width=\linewidth]{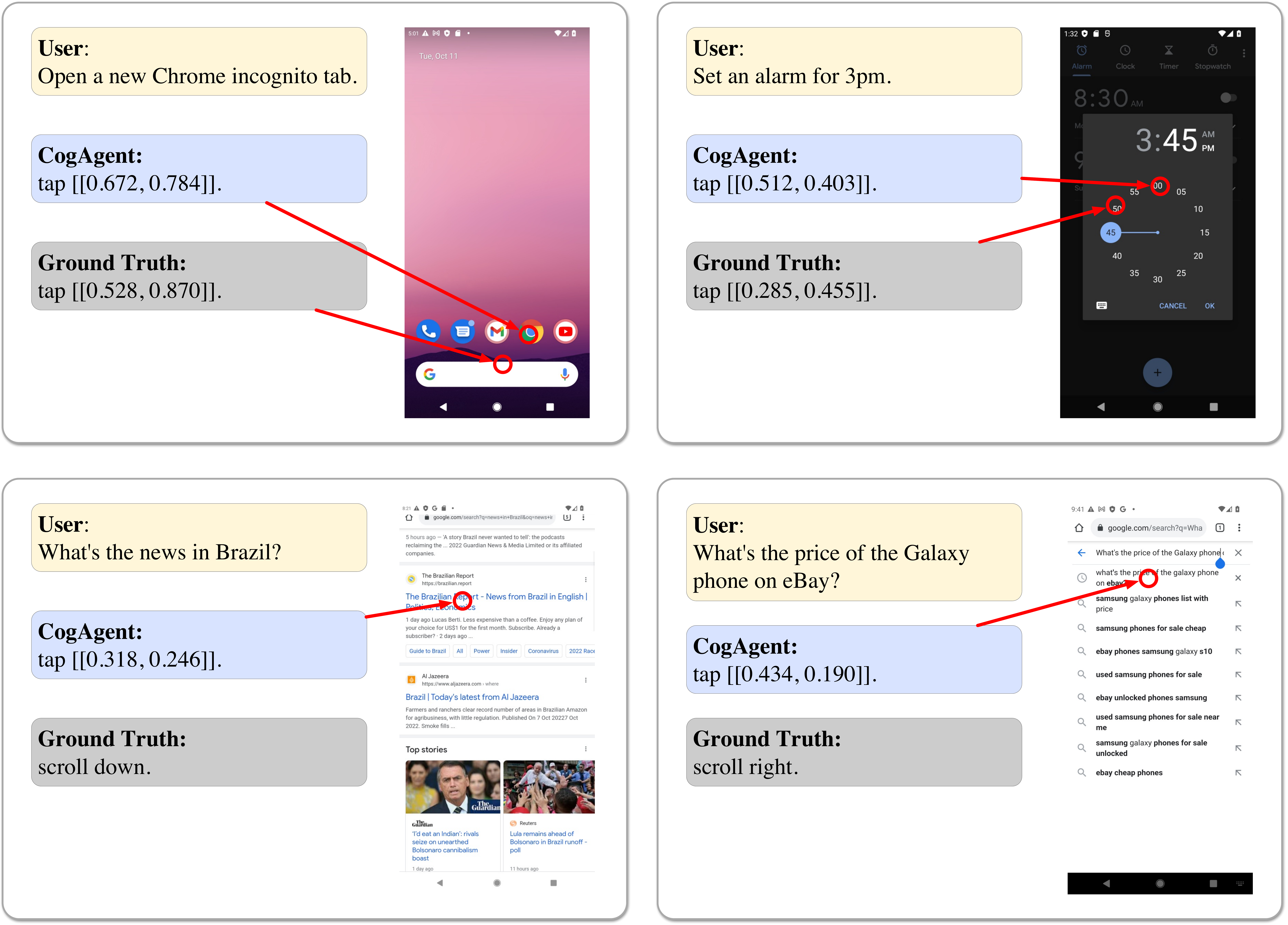}
    \caption{Instances in the AITW dataset where CogAgent provides accurate responses that differ from the annotated ground truth.}
    \label{fig:appendix-analysis-aitw}
    \vspace{10mm}
\end{figure}

In our analysis, we conducted a sampling and manual verification of cases where there was a discrepancy between the model's responses and the labeled answers. We sampled instances and discovered that in 42\% of these, the model's response actually represented an alternative correct method, rather than an error. This phenomenon is common due to the multiplicity of ways to accomplish a function on a smartphone. For example, many phones are equipped with both Google's app and a Google search bar, and searches can be conducted using either. During the use of smartphones, human users may exhibit simultaneous needs for both in-depth exploration and extensive browsing. For example, when browsing news, CogAgent might identify relevant results on the current page and select them immediately, while some users may prefer to scroll through the entire page first. Furthermore, the data collection for annotations might involve redundant or overly detailed steps, whereas CogAgent can directly find the most efficient path. \cref{fig:appendix-analysis-aitw} illustrates these typical examples of multiple correct answers.

Given the nature of smartphone usage, wherein accomplishing a task often involves multiple viable paths, it is advisable for future studies to explore a virtual environment with an evaluation of the accuracy of task completion following a sequence of actions, aligning more closely with real-world human usage scenarios.

% \clearpage
% \begin{strip}
\section{Samples of Pre-train Data}

\begin{figure}[h]
    \centering
    \includegraphics[width=\linewidth]{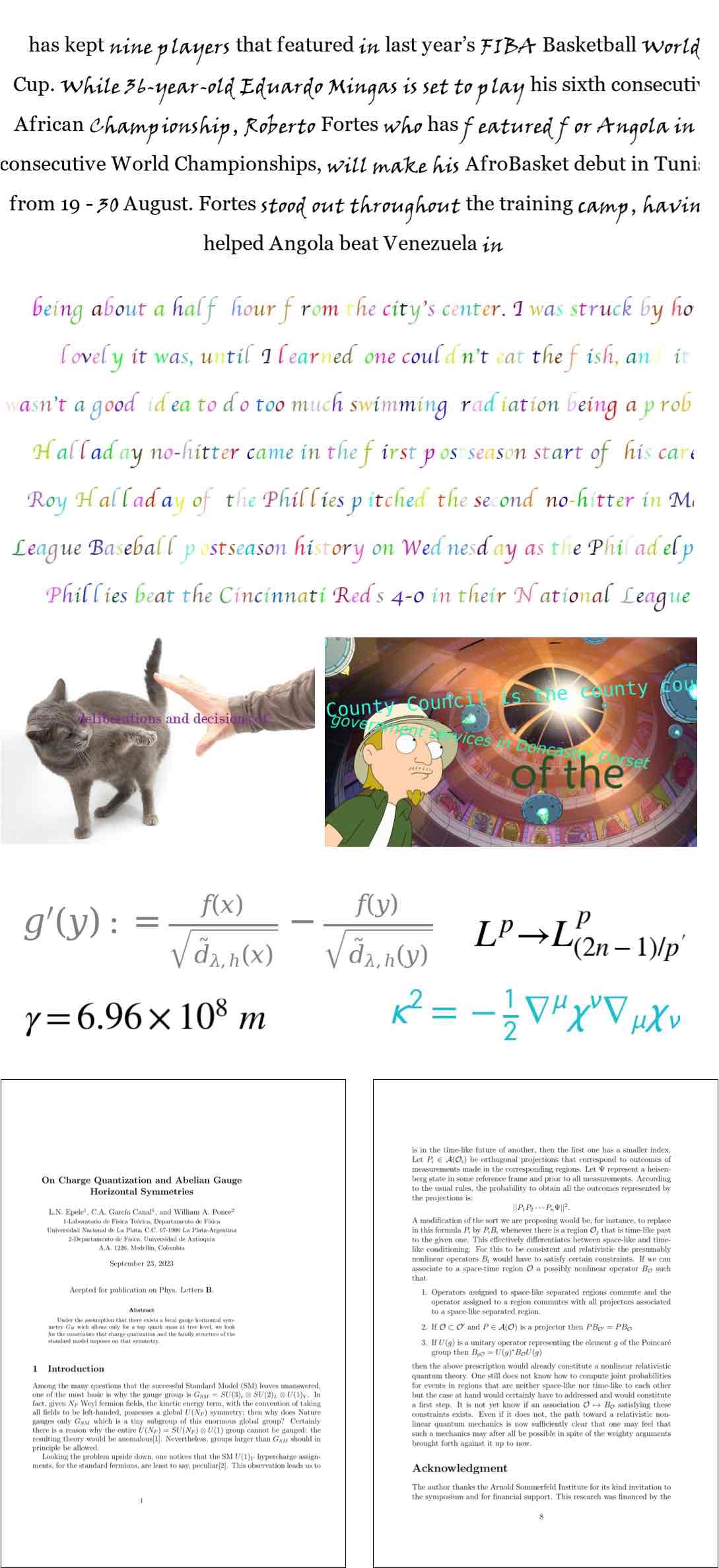}
    \caption{Samples of pre-train data for text recognition.}
    \label{fig:appendix-data-ocr}
\end{figure}

\begin{figure}[b]
    \centering
    \includegraphics[width=0.8\linewidth]{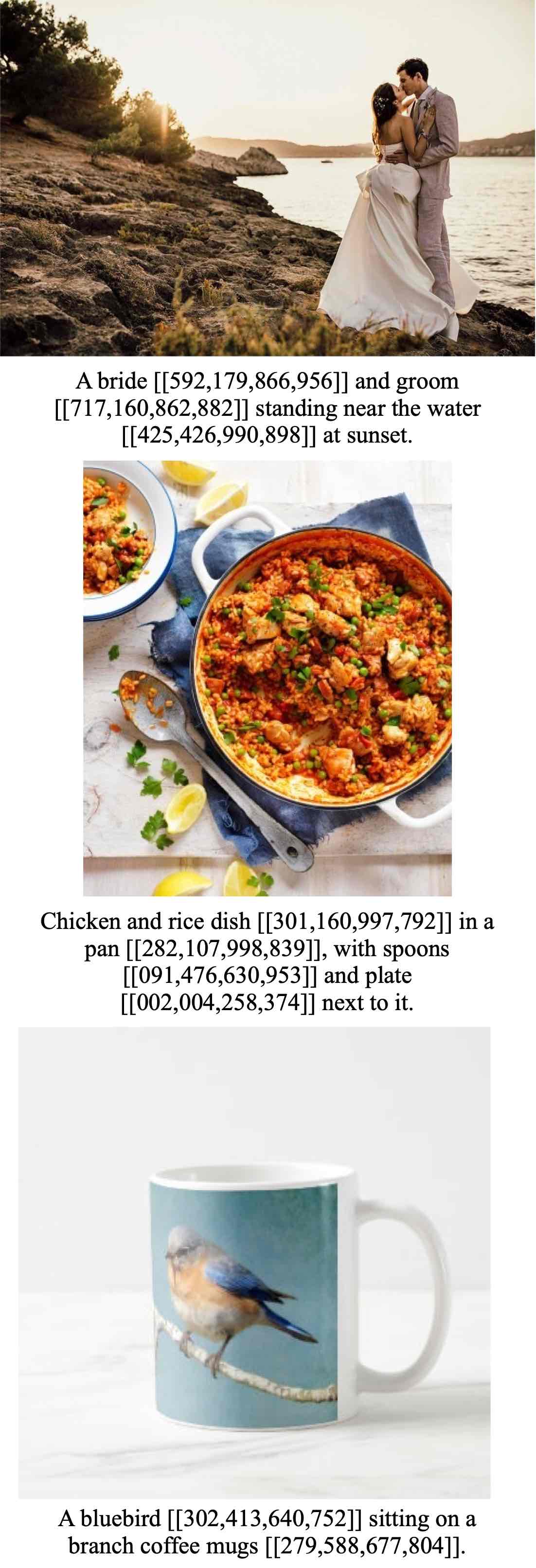}
    \caption{Samples of pre-train data for visual grounding.}
    \label{fig:appendix-data-grounding}
    \vspace{6mm}
\end{figure}

\begin{figure}
    \centering
    \includegraphics[width=\linewidth]{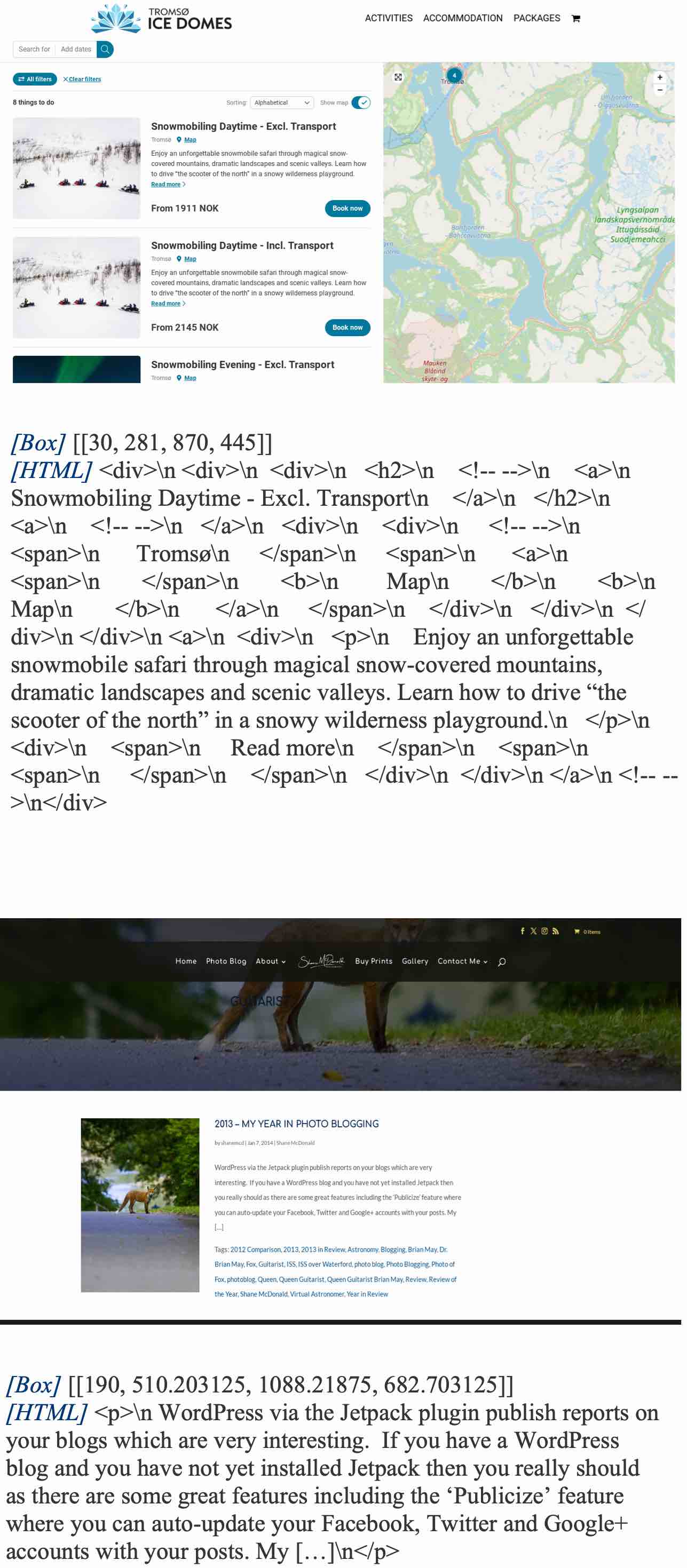}
    \caption{Samples of webpage-html pairs.}
    \label{fig:appendix-data-html}
\end{figure}

Samples of pre-train data are listed in this section, including data for text recognition (\cref{fig:appendix-data-ocr}), visual grounding (\cref{fig:appendix-data-grounding}), and webpage-html pairs (\cref{fig:appendix-data-html}). 

\clearpage

\section{Details of Fine-Tuning Data}

\subsection{Human annotation}
To enable CogAgent to function as an agent on various apps and websites, we assembled a team of over ten annotators to collect and label screenshots. To ensure the quality of the annotations and prevent overly simplistic, homogenized content, the annotation process was divided into two phases.

In the first phase, annotators were required to independently select from a pool of apps and websites, and capture screenshots of pages strongly relevant to the main function of that app/website, with no less than 20 screenshots for each app/website. Subsequently, for each screenshot, the annotators would label them according to the following aspects:
\begin{enumerate}
    \item \textbf{Five buttons.} List the names of five clickable buttons from the screenshots. The types of buttons should be diverse. 
    \item \textbf{Three clickable areas.} List three clickable areas from the screenshots, e.g. text boxes, search boxes, clickable images, hyperlinks, etc.
    \item \textbf{Two questions extracting information from the image.} Ask questions involving the textual information in the images. For instance, ``In what year did the user in the image register?''
    % , which requires information from the user's registration time shown in the image
    .
    \item \textbf{One operation requirement.} What function of the webpage/app would you use in this interface? For example, adding the comment: ``cool!''. The requirement shouldn't be vague such as ``adding comment''.
\end{enumerate}

In the second phase, annotators are required
% to perform grounding annotation for the elements proposed in the first phase (the aforementioned parts 1 and 2), and 
to provide grounding annotation for the questions and operational methods proposed in the first phase (the aforementioned parts 3 and 4). For example, for \Cref{fig:appendix-data-human}, a possible annotation would be:

\begin{figure}[h]
    \centering
    \includegraphics[width=0.7\linewidth]{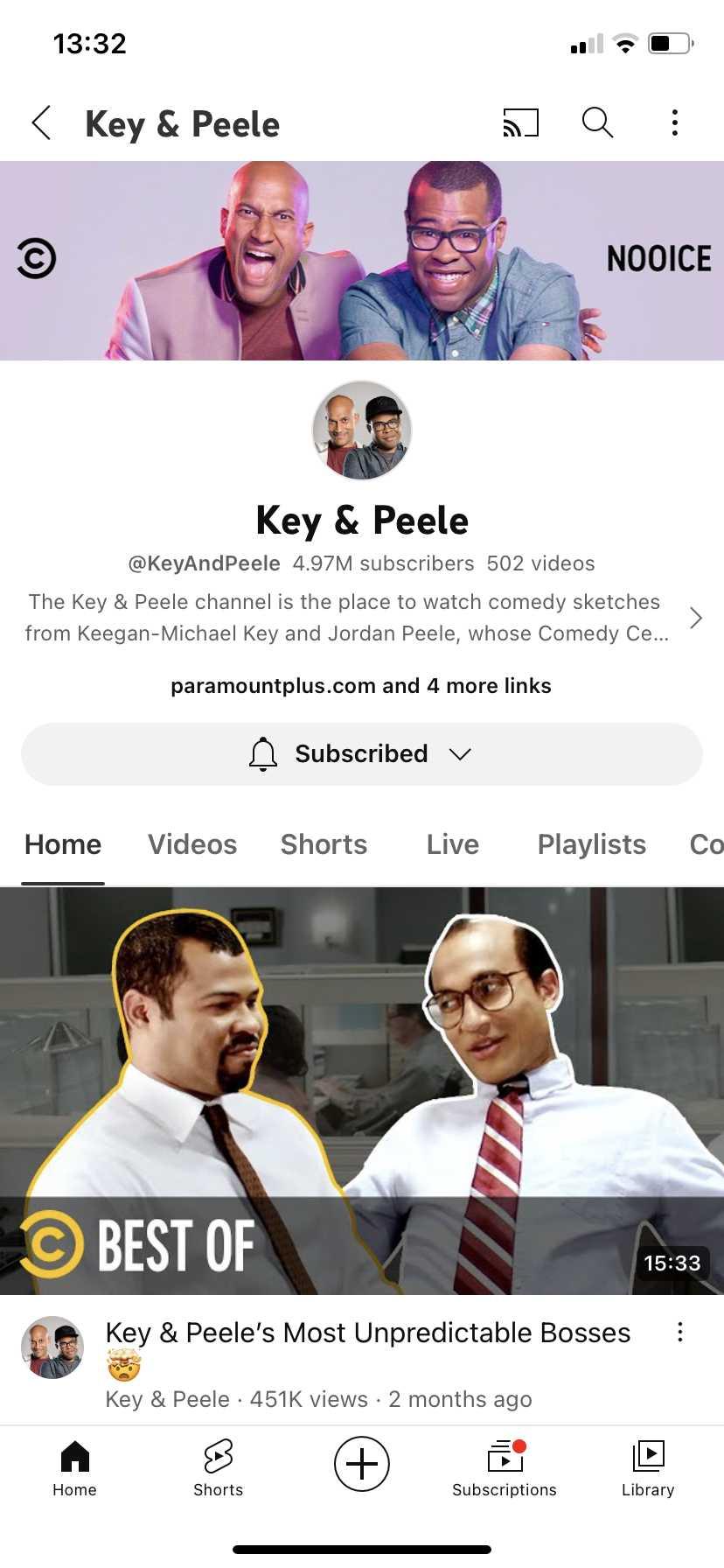}
    \caption{Samples of human-collected screenshot.}
    \label{fig:appendix-data-human}
    \vspace{-3mm}
\end{figure}

\begin{enumerate}
    \item \textbf{Buttons}: Back; Search; Subscribed; Home; Library. 
    \item \textbf{Clickable areas}: Avatar in the middle top; Video preview in the middle bottom; Personal profile.
    \item \textbf{Question 1}: Based on the page, how many followers does this author have? 
    
    \textbf{Answer}: According to the personal profile at the top [[013,568,802,188]], this author has 4.97M followers.

    \textbf{Question 2}: Based on the page, how many videos has this author posted? 
    
    \textbf{Answer}: According to the personal profile at the top [[013,568,802,188]], this author has posted 502 videos.

    \item \textbf{Operation requirement}: Based on the page, write out the steps to complete the following function: Follow this author. 
    
    \textbf{Answer}: Click on Subscribed [[049,826,728,078]] to follow this author.
\end{enumerate}

\subsection{Conversion of Agent Datasets}
% human annotation
To convert Mind2Web to natural language with GPT4, we use the following prompt:

~

\begin{mdframed}[backgroundcolor=gray!10,nobreak]

Imagine that you are a robot operating a computer. Like how humans operate the computer, you can move the mouse, click with the mouse, or type some texts with the keyboard.

**Your ultimate task is: ``Find the lowest-priced round trip flight with hotel on May 2 from Kathmandu, Nepal KTM to Shanghai, China PVG and return on May 5. Book a double room and check out with the default flights.''.**  

\end{mdframed}

\begin{mdframed}[backgroundcolor=gray!10,nobreak]

You are given previous actions: (format: element $\rightarrow$ operation)

% \colorbox{MyBlue}{
1. [link] Flight + Hotel $\rightarrow$ CLICK,

2. [textbox] Where from? $\rightarrow$ TYPE: KATHMANDU,

3. [div] Tribhuvan Intl Airport (KTM), Nepal $\rightarrow$ CLICK,

4. [textbox] Where to? $\rightarrow$ TYPE: SHANGHAI,

5. [div] Pudong Intl Airport (PVG), China $\rightarrow$ CLICK,

6. [span] Sat 8 Apr - Sat 15 Apr $\rightarrow$ CLICK,

7. [checkbox] 2 May 2023 $\rightarrow$ CLICK,

8. [checkbox] 5 May 2023 $\rightarrow$ CLICK,
% }

 Furthermore, you **secretly** know the following steps of action. **(But you have to pretend this is the result of your derivation and you don't know it before)**

 1. [button] Search $\rightarrow$ CLICK,
 
 2. [button] Recommended $\rightarrow$ CLICK,
 
3. [div] Lowest Price $\rightarrow$ CLICK,

4. [svg] $\rightarrow$ CLICK,

5. [button] Choose your room $\rightarrow$ CLICK,

6. [button] Book Business Double Room A $\rightarrow$ CLICK,

7. [svg] $\rightarrow$ CLICK,

8. [svg] $\rightarrow$ CLICK,

9. [button] Choose $\rightarrow$ CLICK

Please formulate an operational guide for future operations for solving the task. The guide includes:

1. Plan: A **multi-step future** plan that does not involve low-level operations **(start from current screen and action, DON'T include previous actions)**; steps indexed by numbers. **Be sure to pretend that you don't know the future interface and actions, and don't know the elements not existing on the current screen.**

2. Action: The specific **immediate** action that needs to be taken (Corresponding to the** very next action** [button] Search $\rightarrow$ CLICK at the box \{``x\_left'': 0.876, ``y\_left'': 0.308, ``width'': 0.063, ``height'': 0.034\}), which should be detailed enough to directly control the operation, such as specifying the exact location (or box) to click, the exact content of the text to type, the item to select.

**Output Format:** A JSON dictionary strictly following the format: 

\{ ``plan'': ``...$<$Your Plan Here$>$'', ``action'': ``...$<$Your Action Here$>$'', ``operation'': ``[button] Search $\rightarrow$ CLICK at the box {``x\_left'': 0.876, ``y\_left'': 0.308, ``width'': 0.063, ``height'': 0.034}''\} 
**Please do not output any content other than the JSON format. Don't mention elements that only appear in HTML such as svg, div.** 
\end{mdframed}

\hide{
\begin{figure}[h]
    \centering
    \includegraphics[width=\linewidth]{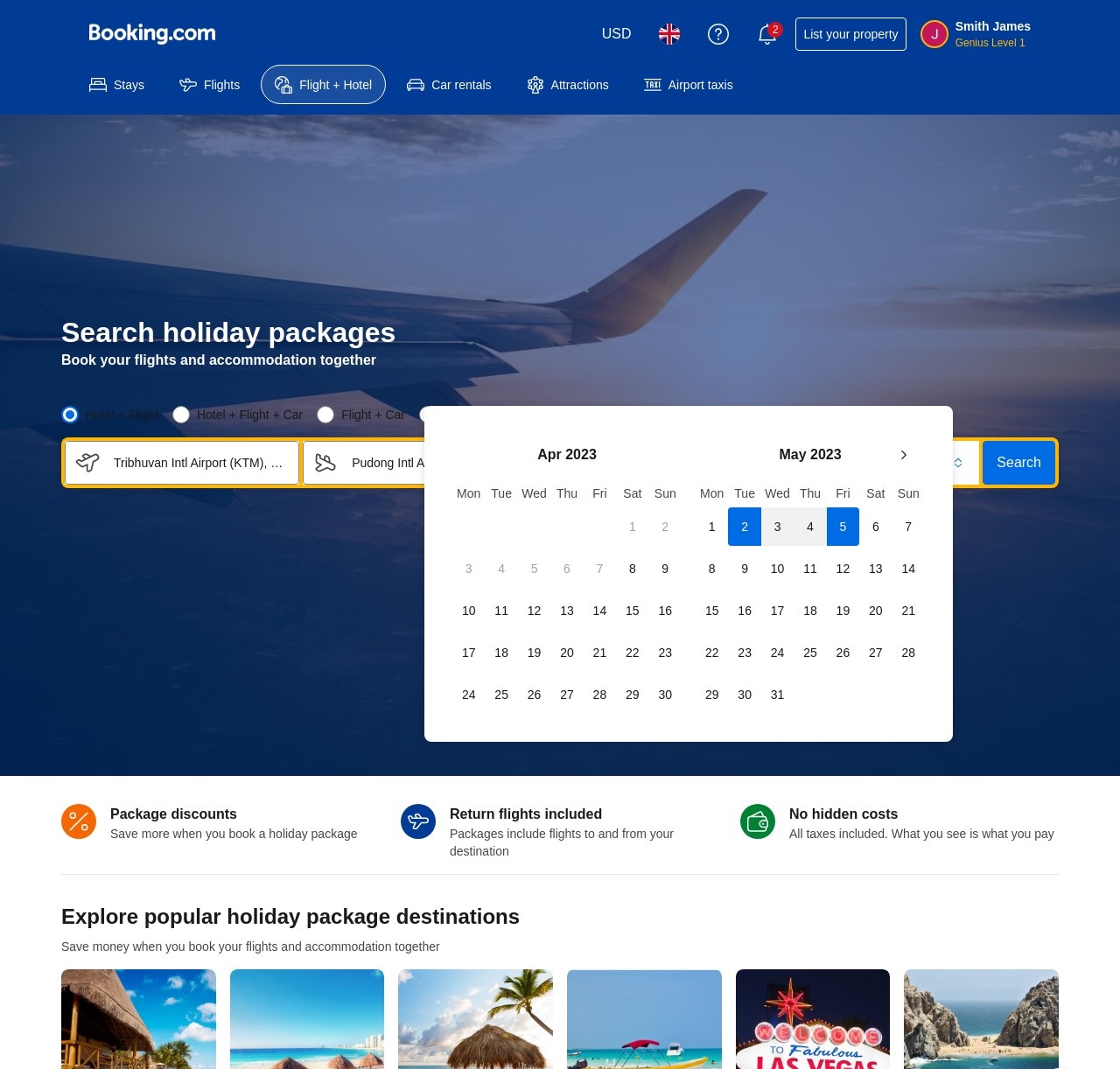}
    \caption{Corresponding screenshot of the prompt in Mind2Web.}
    \label{fig:appendix-data-mind2web}
    \vspace{1mm}
\end{figure}
} %%% end of hide

% \begin{mdframed}[backgroundcolor=gray!10,nobreak]

% **Please do not output any content other than the JSON format. Don't mention elements that only appear in HTML such as SVG, div.** 
% \end{mdframed}

The returned answer from GPT-4 would be:

\begin{mdframed}[backgroundcolor=gray!10,nobreak]
\{
  ``plan'': ``1. After searching, you'll see a list of flight and hotel packages. 2. Review the recommended options, usually highlighted for convenience. 3. Focus on finding the lowest priced option. You might need to navigate through various sorting or filtering features to do this. 4. Once you've found the lowest priced option, select it to view more details. 5. In the detailed view, you'll have to choose your room type. Look for options that fit your requirements, like a business double room. 6. After selecting the room, proceed to finalize your booking. This might involve confirming your choice and entering any required information. 7. Review all the details carefully before finalizing the booking to ensure everything is correct.'', 
  
  ``action'': ``Click the 'Search' button to proceed with the search based on the entered criteria. This button is located towards the right side of the screen.'', 
  
  ``operation'': ``[button] Search $\rightarrow$  CLICK at the box \{``x\_left'': 0.876, ``y\_left'': 0.308, ``width'': 0.063, ``height'': 0.03\}''
\}
\end{mdframed}

% \end{strip}
% \clearpage

\section{Failure cases}
\begin{figure}[h]
    \centering
    \includegraphics[width=\linewidth]{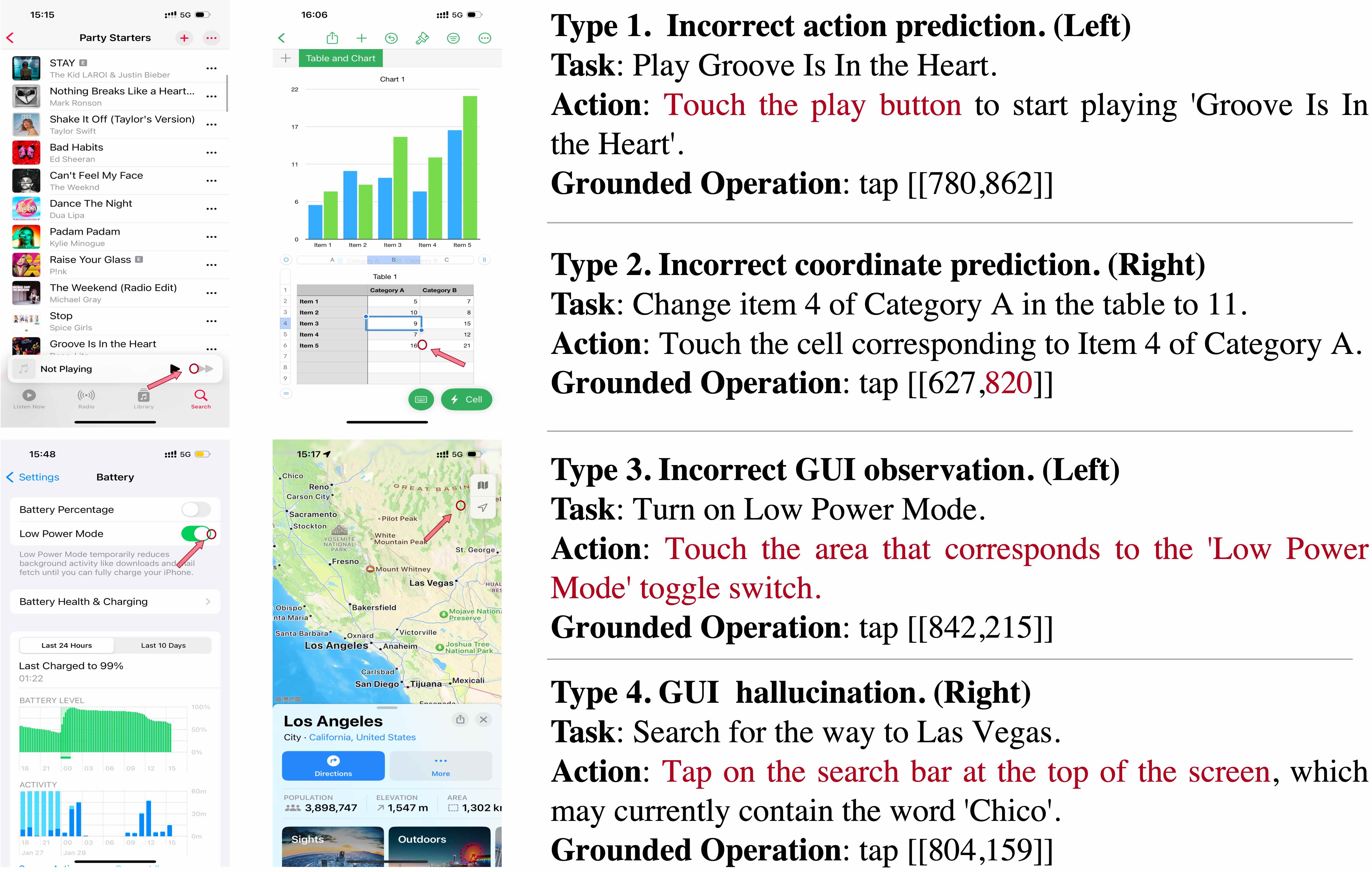}
    \caption{Failure cases of GUI Agent. }
    \label{fig:failure_case}
\end{figure}

The failure cases can be summarized into four categories (\cref{fig:failure_case}): 
\begin{enumerate}
    \item Incorrect action prediction;
    \item Incorrect coordinate prediction;
    \item Incorrect GUI observation;
    \item Hallucination.
\end{enumerate}

\section{More Generated Samples of CogAgent}
To further demonstrate the capabilities of CogAgent, we provide additional generated samples in this section. 

% \begin{figure}[h]
%     \centering
%     \includegraphics[width=\linewidth]{figures/appendix/appendix_ppt.jpg}
%     \caption{More generated samples of CogAgent.}
%     \label{fig:appendix-data-mind2web}
%     % \vspace{-5mm}
% \end{figure}

\begin{figure*}[bp]
    \centering
    \includegraphics[width=0.9\linewidth]{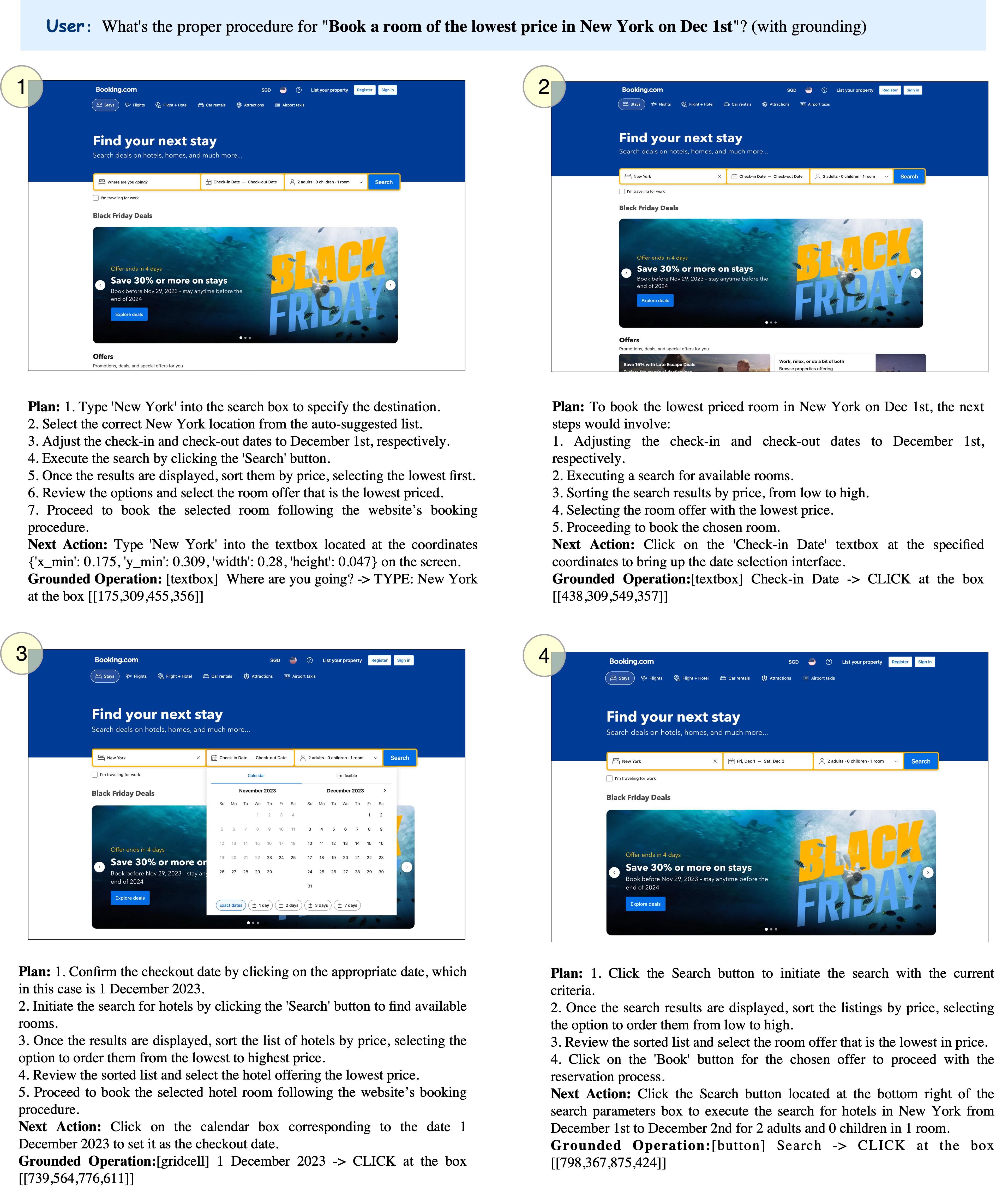}
    \caption{More generated samples of CogAgent.}
\end{figure*}

\begin{figure*}
    \centering
    \includegraphics[width=0.9\linewidth]{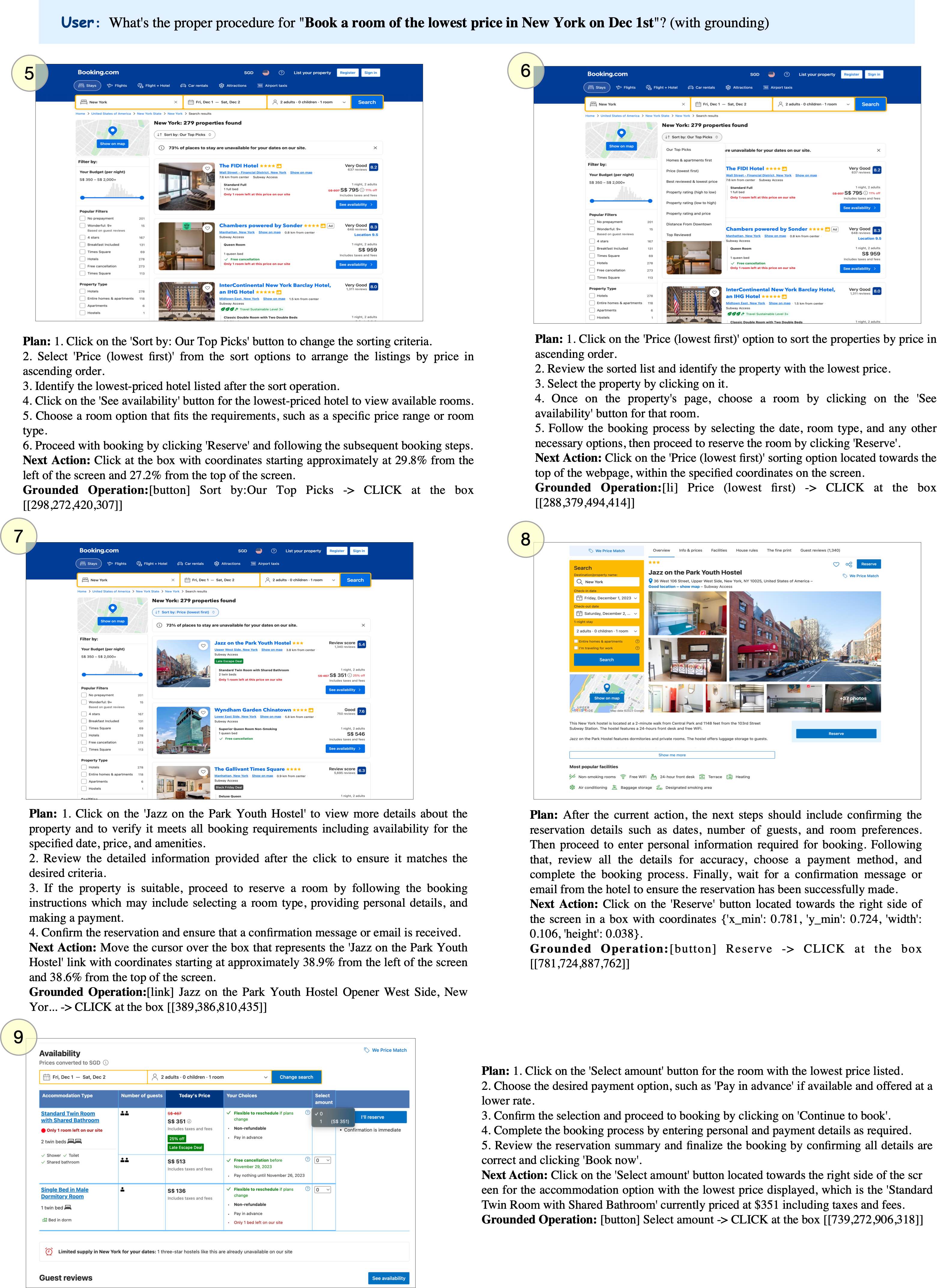}
    \caption{More generated samples of CogAgent.}
\end{figure*}

\begin{figure*}
    \centering
    \includegraphics[width=0.9\linewidth]{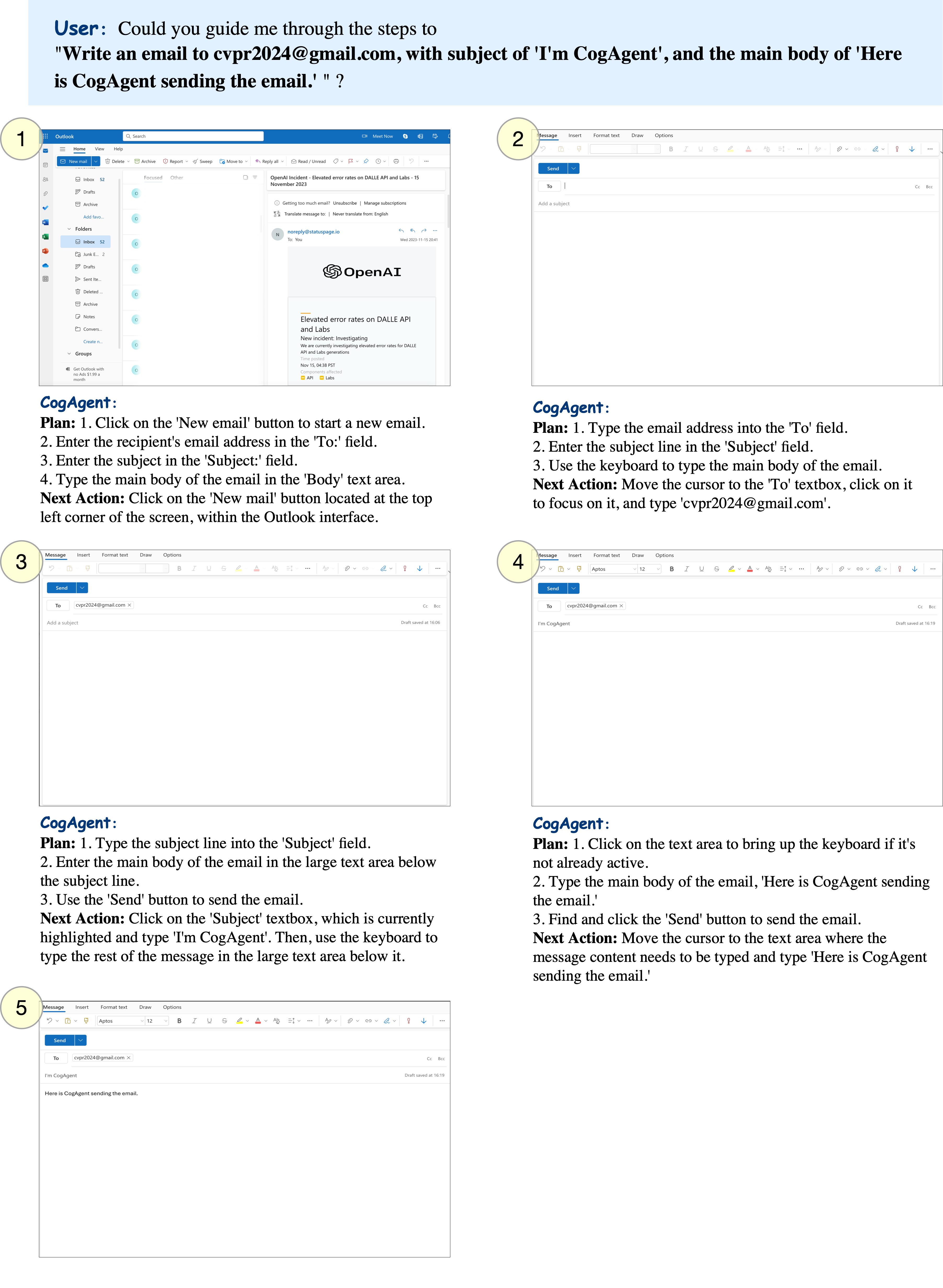}
    \caption{More generated samples of CogAgent.}
\end{figure*}

\begin{figure*}
    \centering
    \includegraphics[width=0.9\linewidth]{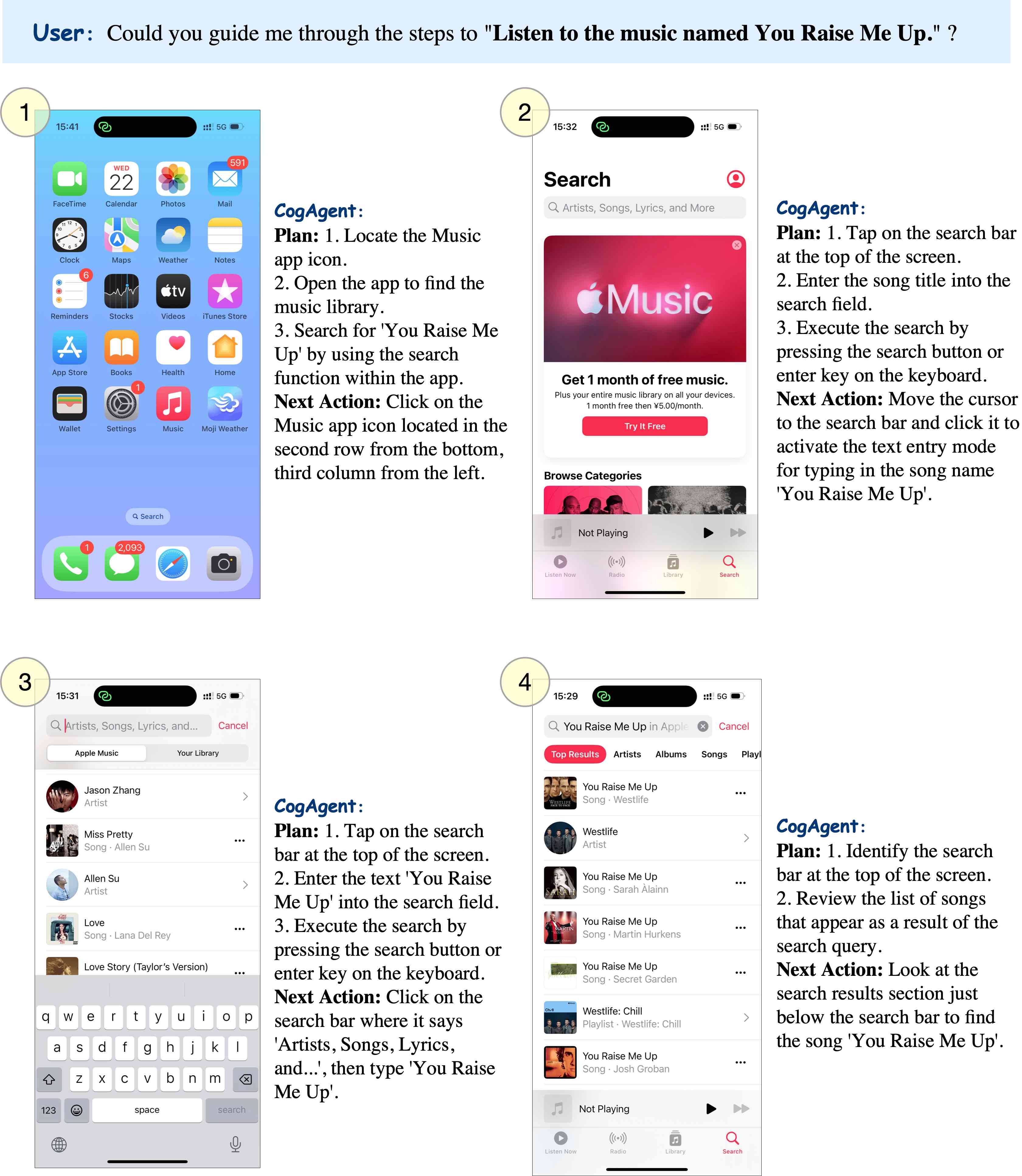}
    \vspace{6mm}
    \caption{More generated samples of CogAgent.}
\end{figure*}

\begin{figure*}
    \centering
    \includegraphics[width=0.8\linewidth]{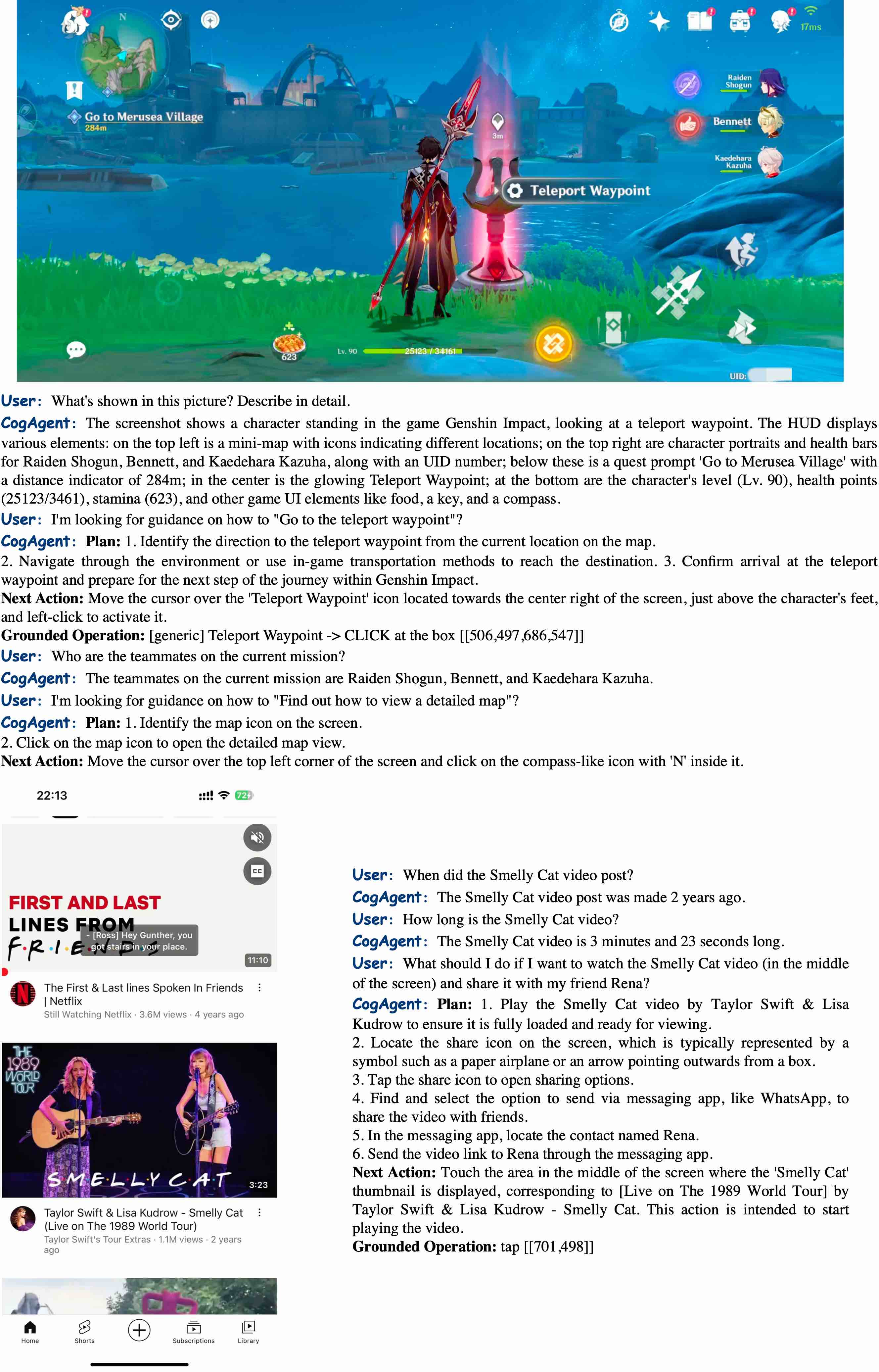}
    \vspace{-3mm}
    \caption{More generated samples of CogAgent.}
\end{figure*}

\begin{figure*}
    \centering
    \includegraphics[width=0.8\linewidth]{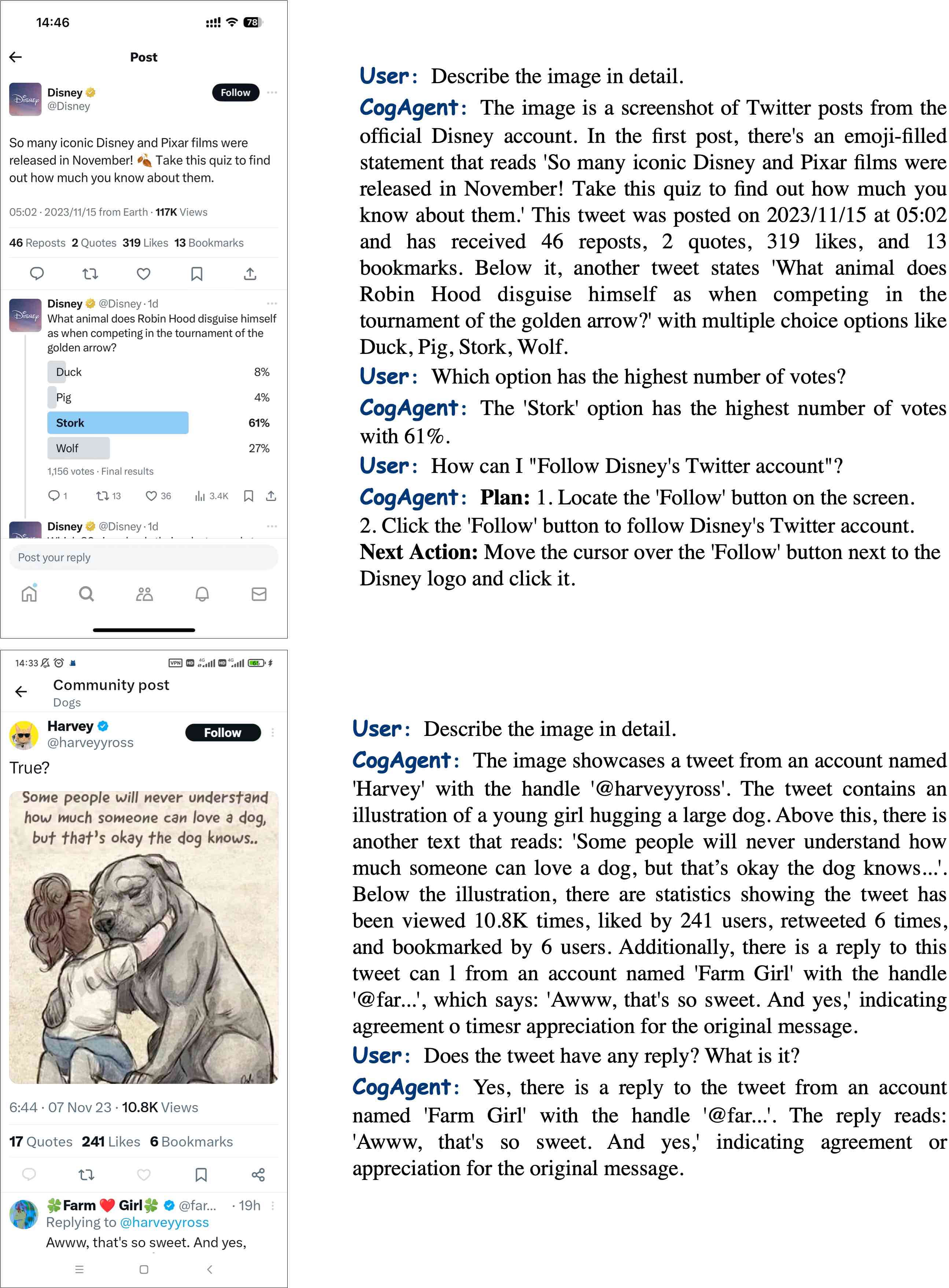}
    \caption{More generated samples of CogAgent.}
\end{figure*}

\begin{figure*}
    \centering
    \includegraphics[width=\linewidth]{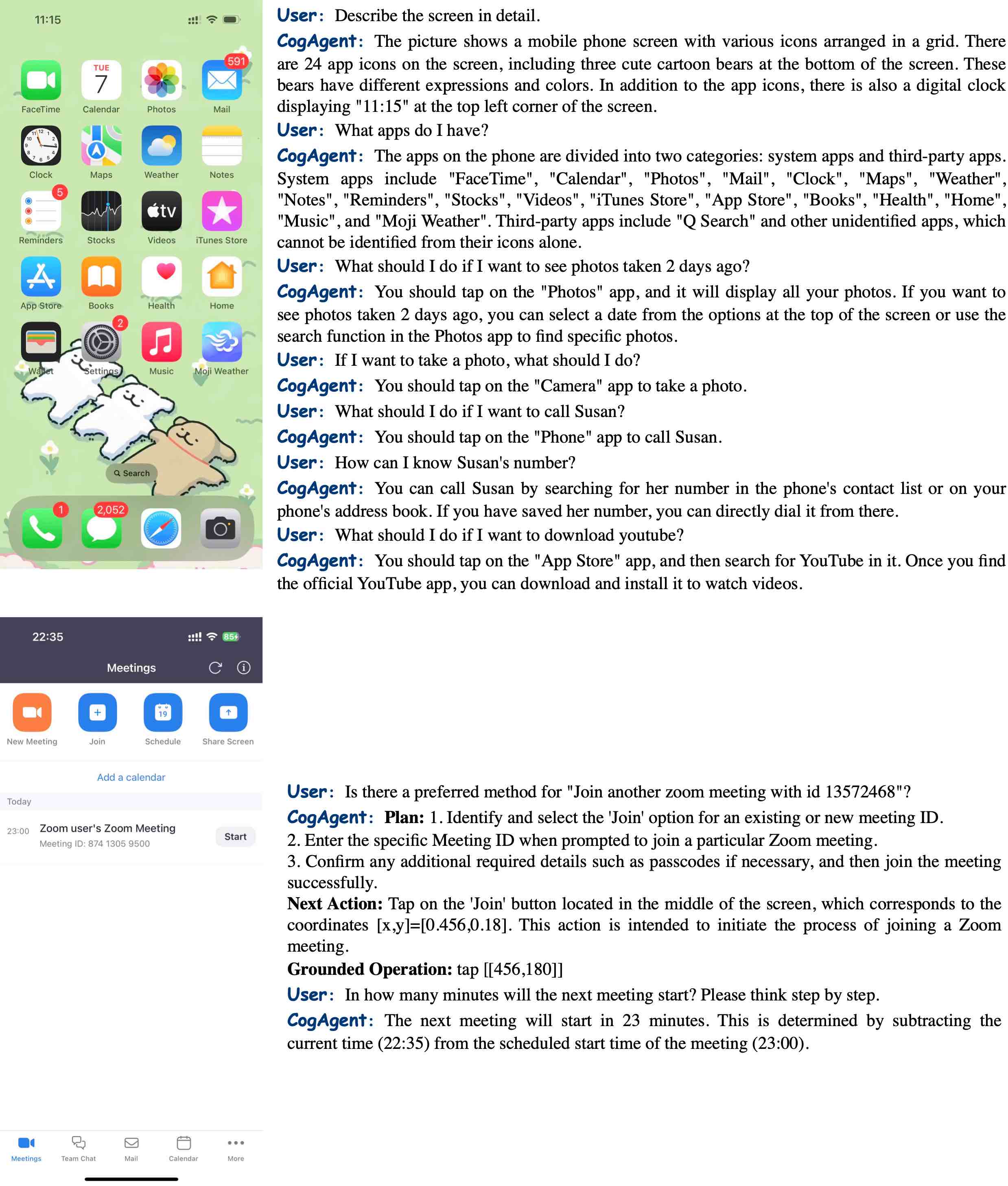}
    \caption{More generated samples of CogAgent.}
\end{figure*}

\begin{figure*}
    \centering
    \includegraphics[width=0.8\linewidth]{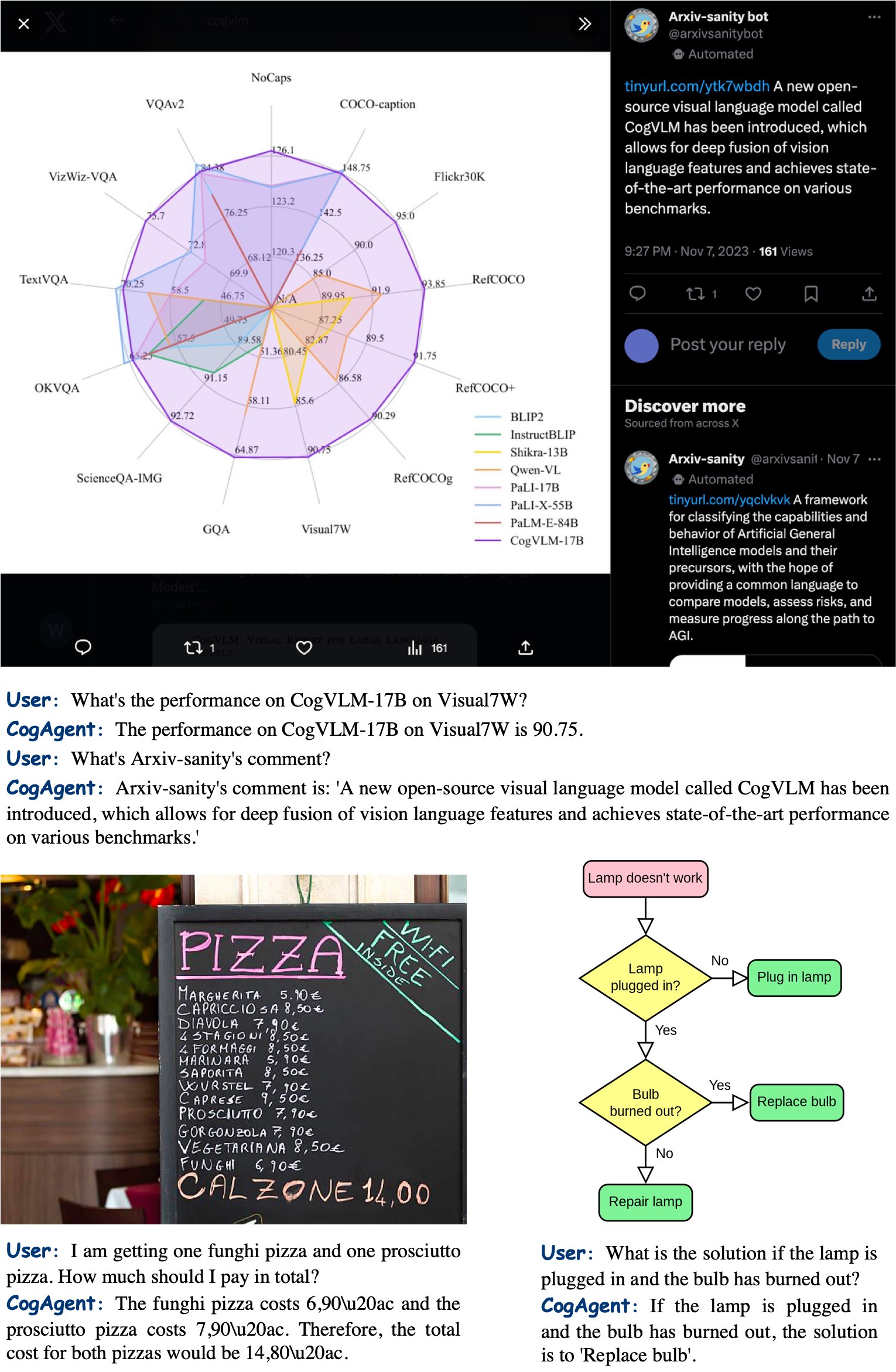}
    \caption{More generated samples of CogAgent.}
\end{figure*}

\begin{figure*}
    \centering
    \includegraphics[width=0.8\linewidth]{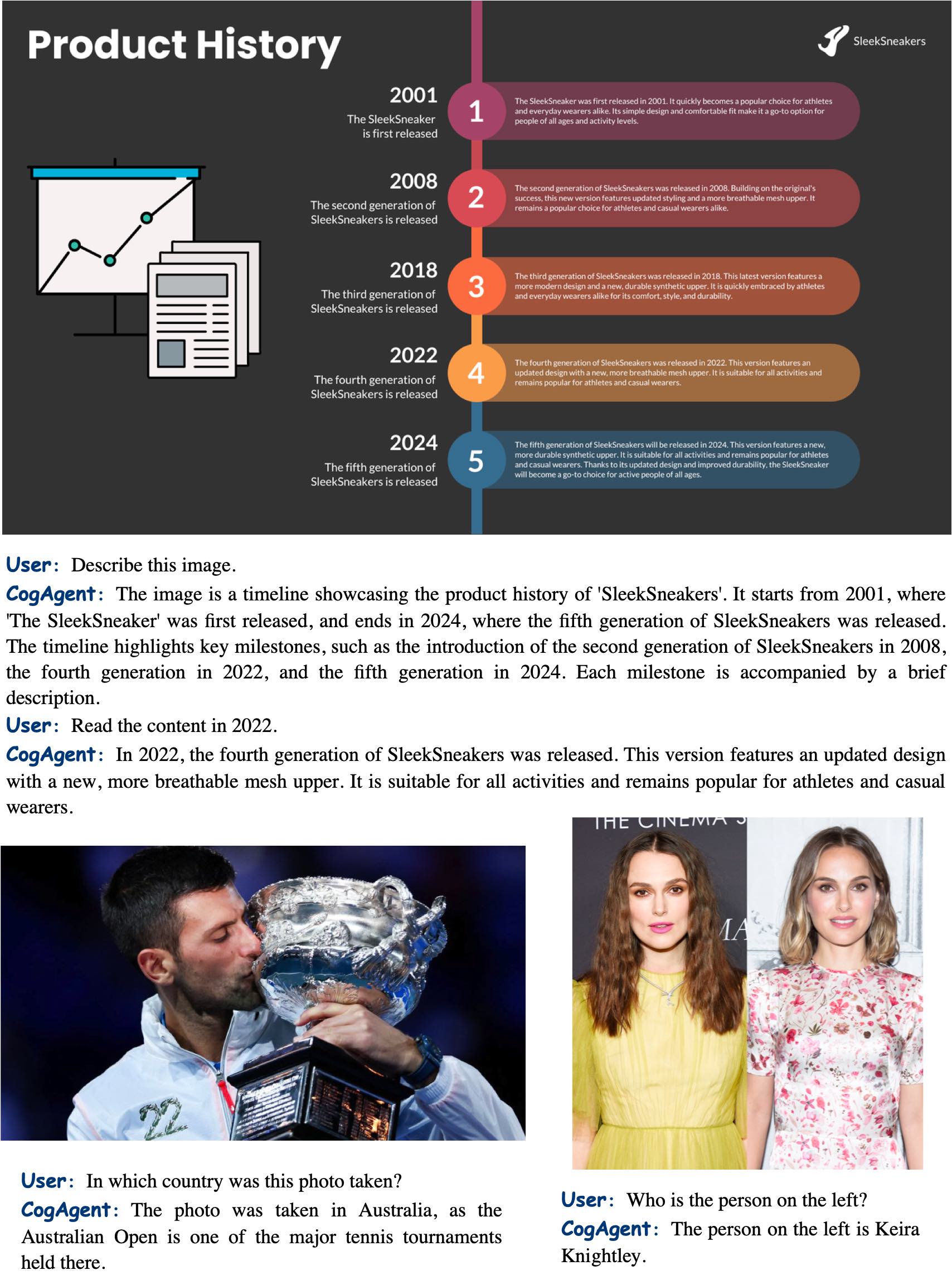}
    \caption{More generated samples of CogAgent.}
\end{figure*}

\begin{figure*}
    \centering
    \includegraphics[width=\linewidth]{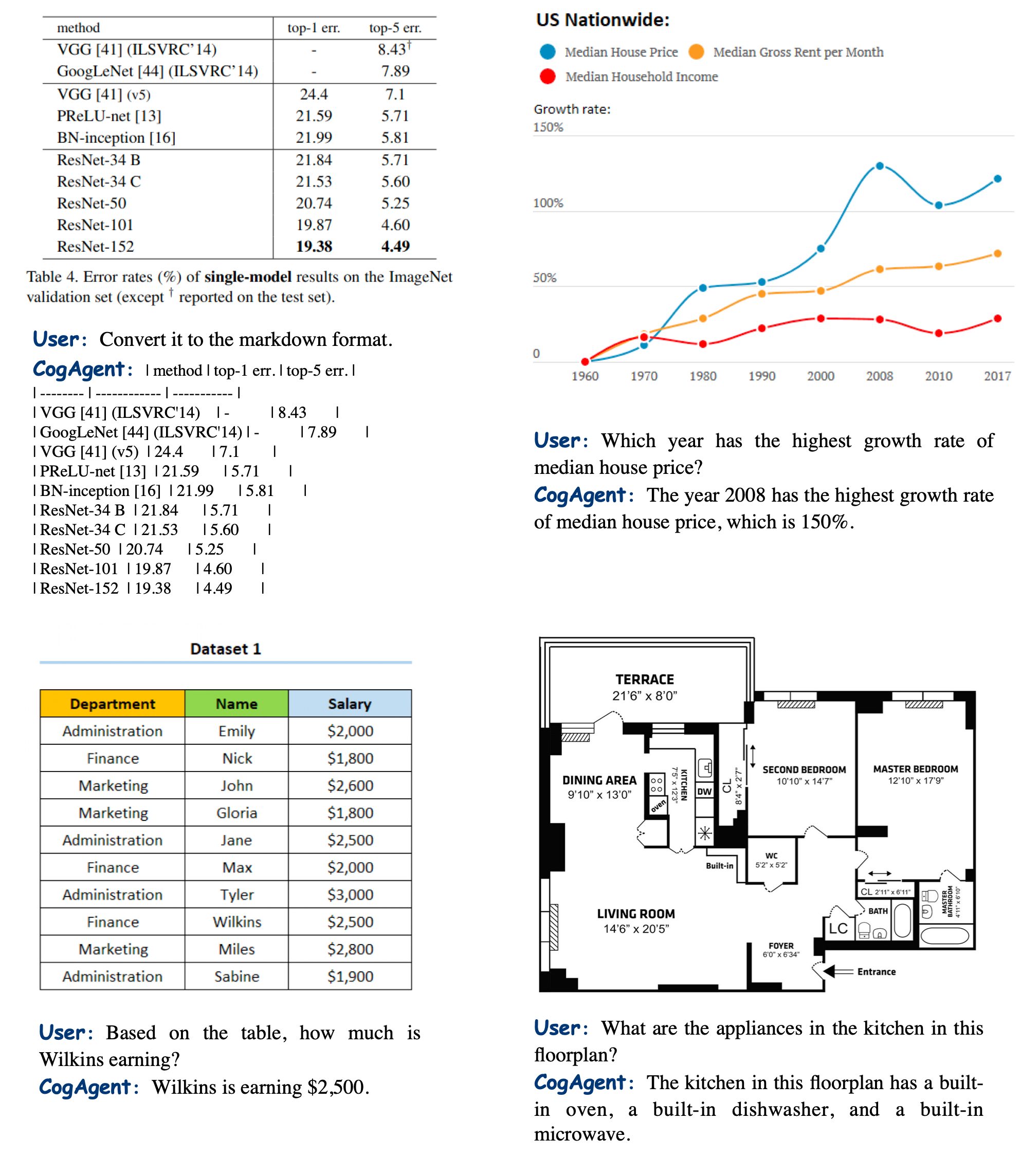}
    \caption{More generated samples of CogAgent.}
\end{figure*}

\hide{ %start of hide

 \clearpage
 {
     \small
     \bibliographystyle{ieeenat_fullname}
     \bibliography{main}

\begin{thebibliography}{44}
\providecommand{\natexlab}[1]{#1}
\providecommand{\url}[1]{\texttt{#1}}
\expandafter\ifx\csname urlstyle\endcsname\relax
  \providecommand{\doi}[1]{doi: #1}\else
  \providecommand{\doi}{doi: \begingroup \urlstyle{rm}\Url}\fi

\bibitem[Antol et~al.(2015)Antol, Agrawal, Lu, Mitchell, Batra, Zitnick, and Parikh]{antol2015vqa}
Stanislaw Antol, Aishwarya Agrawal, Jiasen Lu, Margaret Mitchell, Dhruv Batra, C~Lawrence Zitnick, and Devi Parikh.
\newblock Vqa: Visual question answering.
\newblock In \emph{Proceedings of the IEEE international conference on computer vision}, pages 2425--2433, 2015.

\bibitem[Bai et~al.(2023)Bai, Bai, Yang, Wang, Tan, Wang, Lin, Zhou, and Zhou]{bai2023qwen}
Jinze Bai, Shuai Bai, Shusheng Yang, Shijie Wang, Sinan Tan, Peng Wang, Junyang Lin, Chang Zhou, and Jingren Zhou.
\newblock Qwen-vl: A frontier large vision-language model with versatile abilities.
\newblock \emph{arXiv preprint arXiv:2308.12966}, 2023.

\bibitem[Bavishi et~al.(2023)Bavishi, Elsen, Hawthorne, Nye, Odena, Somani, and Ta\c{s}\i{}rlar]{fuyu-8b}
Rohan Bavishi, Erich Elsen, Curtis Hawthorne, Maxwell Nye, Augustus Odena, Arushi Somani, and Sa\u{g}nak Ta\c{s}\i{}rlar.
\newblock Introducing our multimodal models, 2023.

\bibitem[Biten et~al.(2019)Biten, Tito, Mafla, Gomez, Rusinol, Valveny, Jawahar, and Karatzas]{biten2019scene}
Ali~Furkan Biten, Ruben Tito, Andres Mafla, Lluis Gomez, Mar{\c{c}}al Rusinol, Ernest Valveny, CV Jawahar, and Dimosthenis Karatzas.
\newblock Scene text visual question answering.
\newblock In \emph{Proceedings of the IEEE/CVF international conference on computer vision}, pages 4291--4301, 2019.

\bibitem[Blecher et~al.(2023)Blecher, Cucurull, Scialom, and Stojnic]{blecher2023nougat}
Lukas Blecher, Guillem Cucurull, Thomas Scialom, and Robert Stojnic.
\newblock Nougat: Neural optical understanding for academic documents.
\newblock \emph{arXiv preprint arXiv:2308.13418}, 2023.

\bibitem[Byeon et~al.(2022)Byeon, Park, Kim, Lee, Baek, and Kim]{kakaobrain2022coyo-700m}
Minwoo Byeon, Beomhee Park, Haecheon Kim, Sungjun Lee, Woonhyuk Baek, and Saehoon Kim.
\newblock Coyo-700m: Image-text pair dataset.
\newblock \url{https://github.com/kakaobrain/coyo-dataset}, 2022.

\bibitem[Chen et~al.(2023{\natexlab{a}})Chen, Shu, Shareghi, Collier, Narasimhan, and Yao]{chen2023fireact}
Baian Chen, Chang Shu, Ehsan Shareghi, Nigel Collier, Karthik Narasimhan, and Shunyu Yao.
\newblock Fireact: Toward language agent fine-tuning.
\newblock \emph{arXiv preprint arXiv:2310.05915}, 2023{\natexlab{a}}.

\bibitem[Chen et~al.(2023{\natexlab{b}})Chen, Djolonga, Padlewski, Mustafa, Changpinyo, Wu, Ruiz, Goodman, Wang, Tay, et~al.]{chen2023pali}
Xi Chen, Josip Djolonga, Piotr Padlewski, Basil Mustafa, Soravit Changpinyo, Jialin Wu, Carlos~Riquelme Ruiz, Sebastian Goodman, Xiao Wang, Yi Tay, et~al.
\newblock Pali-x: On scaling up a multilingual vision and language model.
\newblock \emph{arXiv preprint arXiv:2305.18565}, 2023{\natexlab{b}}.

\bibitem[Dai et~al.(2023)Dai, Li, Li, Tiong, Zhao, Wang, Li, Fung, and Hoi]{dai2023instructblip}
Wenliang Dai, Junnan Li, Dongxu Li, Anthony Meng~Huat Tiong, Junqi Zhao, Weisheng Wang, Boyang Li, Pascale Fung, and Steven Hoi.
\newblock Instructblip: Towards general-purpose vision-language models with instruction tuning, 2023.

\bibitem[Deng et~al.(2023)Deng, Gu, Zheng, Chen, Stevens, Wang, Sun, and Su]{deng2023mind2web}
Xiang Deng, Yu Gu, Boyuan Zheng, Shijie Chen, Samuel Stevens, Boshi Wang, Huan Sun, and Yu Su.
\newblock Mind2web: Towards a generalist agent for the web.
\newblock \emph{arXiv preprint arXiv:2306.06070}, 2023.

\bibitem[Dong et~al.(2023)Dong, Han, Peng, Qi, Ge, Yang, Zhao, Sun, Zhou, Wei, et~al.]{dong2023dreamllm}
Runpei Dong, Chunrui Han, Yuang Peng, Zekun Qi, Zheng Ge, Jinrong Yang, Liang Zhao, Jianjian Sun, Hongyu Zhou, Haoran Wei, et~al.
\newblock Dreamllm: Synergistic multimodal comprehension and creation.
\newblock \emph{arXiv preprint arXiv:2309.11499}, 2023.

\bibitem[Dosovitskiy et~al.(2020)Dosovitskiy, Beyer, Kolesnikov, Weissenborn, Zhai, Unterthiner, Dehghani, Minderer, Heigold, Gelly, et~al.]{dosovitskiy2020image}
Alexey Dosovitskiy, Lucas Beyer, Alexander Kolesnikov, Dirk Weissenborn, Xiaohua Zhai, Thomas Unterthiner, Mostafa Dehghani, Matthias Minderer, Georg Heigold, Sylvain Gelly, et~al.
\newblock An image is worth 16x16 words: Transformers for image recognition at scale.
\newblock \emph{arXiv preprint arXiv:2010.11929}, 2020.

\bibitem[Du et~al.(2020)Du, Li, Guo, Yin, Liu, Zhou, Bai, Yu, Yang, Dang, et~al.]{du2020pp}
Yuning Du, Chenxia Li, Ruoyu Guo, Xiaoting Yin, Weiwei Liu, Jun Zhou, Yifan Bai, Zilin Yu, Yehua Yang, Qingqing Dang, et~al.
\newblock Pp-ocr: A practical ultra lightweight ocr system.
\newblock \emph{arXiv preprint arXiv:2009.09941}, 2020.

\bibitem[Gao et~al.(2023)Gao, Han, Zhang, Lin, Geng, Zhou, Zhang, Lu, He, Yue, et~al.]{gao2023llamaadp}
Peng Gao, Jiaming Han, Renrui Zhang, Ziyi Lin, Shijie Geng, Aojun Zhou, Wei Zhang, Pan Lu, Conghui He, Xiangyu Yue, et~al.
\newblock Llama-adapter v2: Parameter-efficient visual instruction model.
\newblock \emph{arXiv preprint arXiv:2304.15010}, 2023.

\bibitem[Kim et~al.(2022)Kim, Hong, Yim, Nam, Park, Yim, Hwang, Yun, Han, and Park]{kim2022ocr}
Geewook Kim, Teakgyu Hong, Moonbin Yim, JeongYeon Nam, Jinyoung Park, Jinyeong Yim, Wonseok Hwang, Sangdoo Yun, Dongyoon Han, and Seunghyun Park.
\newblock Ocr-free document understanding transformer.
\newblock In \emph{European Conference on Computer Vision}, pages 498--517. Springer, 2022.

\bibitem[Lee et~al.(2023)Lee, Joshi, Turc, Hu, Liu, Eisenschlos, Khandelwal, Shaw, Chang, and Toutanova]{lee2023pix2struct}
Kenton Lee, Mandar Joshi, Iulia~Raluca Turc, Hexiang Hu, Fangyu Liu, Julian~Martin Eisenschlos, Urvashi Khandelwal, Peter Shaw, Ming-Wei Chang, and Kristina Toutanova.
\newblock Pix2struct: Screenshot parsing as pretraining for visual language understanding.
\newblock In \emph{International Conference on Machine Learning}, pages 18893--18912. PMLR, 2023.

\bibitem[Li et~al.(2023{\natexlab{a}})Li, Zhang, Chen, Wang, Pu, Yang, Li, and Liu]{li2023mimic}
Bo Li, Yuanhan Zhang, Liangyu Chen, Jinghao Wang, Fanyi Pu, Jingkang Yang, Chunyuan Li, and Ziwei Liu.
\newblock Mimic-it: Multi-modal in-context instruction tuning.
\newblock \emph{arXiv preprint arXiv:2306.05425}, 2023{\natexlab{a}}.

\bibitem[Li et~al.(2023{\natexlab{b}})Li, Li, Savarese, and Hoi]{li2023blip}
Junnan Li, Dongxu Li, Silvio Savarese, and Steven Hoi.
\newblock Blip-2: Bootstrapping language-image pre-training with frozen image encoders and large language models.
\newblock \emph{arXiv preprint arXiv:2301.12597}, 2023{\natexlab{b}}.

\bibitem[Li et~al.(2023{\natexlab{c}})Li, Du, Zhou, Wang, Zhao, and Wen]{li2023evaluating}
Yifan Li, Yifan Du, Kun Zhou, Jinpeng Wang, Wayne~Xin Zhao, and Ji-Rong Wen.
\newblock Evaluating object hallucination in large vision-language models.
\newblock \emph{arXiv preprint arXiv:2305.10355}, 2023{\natexlab{c}}.

\bibitem[Liu et~al.(2023{\natexlab{a}})Liu, Li, Li, and Lee]{liu2023improved}
Haotian Liu, Chunyuan Li, Yuheng Li, and Yong~Jae Lee.
\newblock Improved baselines with visual instruction tuning.
\newblock \emph{arXiv preprint arXiv:2310.03744}, 2023{\natexlab{a}}.

\bibitem[Liu et~al.(2023{\natexlab{b}})Liu, Li, Wu, and Lee]{liu2023visual}
Haotian Liu, Chunyuan Li, Qingyang Wu, and Yong~Jae Lee.
\newblock Visual instruction tuning.
\newblock \emph{arXiv preprint arXiv:2304.08485}, 2023{\natexlab{b}}.

\bibitem[Lv et~al.(2023)Lv, Huang, Chen, Cui, Ma, Chang, Huang, Wang, Dong, Luo, et~al.]{lv2023kosmos}
Tengchao Lv, Yupan Huang, Jingye Chen, Lei Cui, Shuming Ma, Yaoyao Chang, Shaohan Huang, Wenhui Wang, Li Dong, Weiyao Luo, et~al.
\newblock Kosmos-2.5: A multimodal literate model.
\newblock \emph{arXiv preprint arXiv:2309.11419}, 2023.

\bibitem[Marino et~al.(2019)Marino, Rastegari, Farhadi, and Mottaghi]{marino2019ok}
Kenneth Marino, Mohammad Rastegari, Ali Farhadi, and Roozbeh Mottaghi.
\newblock Ok-vqa: A visual question answering benchmark requiring external knowledge.
\newblock In \emph{Proceedings of the IEEE/cvf conference on computer vision and pattern recognition}, pages 3195--3204, 2019.

\bibitem[Masry et~al.(2022)Masry, Long, Tan, Joty, and Hoque]{masry2022chartqa}
Ahmed Masry, Do~Xuan Long, Jia~Qing Tan, Shafiq Joty, and Enamul Hoque.
\newblock Chartqa: A benchmark for question answering about charts with visual and logical reasoning.
\newblock \emph{arXiv preprint arXiv:2203.10244}, 2022.

\bibitem[Mathew et~al.(2021)Mathew, Karatzas, and Jawahar]{mathew2021docvqa}
Minesh Mathew, Dimosthenis Karatzas, and CV Jawahar.
\newblock Docvqa: A dataset for vqa on document images.
\newblock In \emph{Proceedings of the IEEE/CVF winter conference on applications of computer vision}, pages 2200--2209, 2021.

\bibitem[Mathew et~al.(2022)Mathew, Bagal, Tito, Karatzas, Valveny, and Jawahar]{mathew2022infographicvqa}
Minesh Mathew, Viraj Bagal, Rub{\`e}n Tito, Dimosthenis Karatzas, Ernest Valveny, and CV Jawahar.
\newblock Infographicvqa.
\newblock In \emph{Proceedings of the IEEE/CVF Winter Conference on Applications of Computer Vision}, pages 1697--1706, 2022.

\bibitem[Mishra et~al.(2019)Mishra, Shekhar, Singh, and Chakraborty]{mishra2019ocr}
Anand Mishra, Shashank Shekhar, Ajeet~Kumar Singh, and Anirban Chakraborty.
\newblock Ocr-vqa: Visual question answering by reading text in images.
\newblock In \emph{2019 international conference on document analysis and recognition (ICDAR)}, pages 947--952. IEEE, 2019.

\bibitem[Nakano et~al.(2021)Nakano, Hilton, Balaji, Wu, Ouyang, Kim, Hesse, Jain, Kosaraju, Saunders, et~al.]{nakano2021webgpt}
Reiichiro Nakano, Jacob Hilton, Suchir Balaji, Jeff Wu, Long Ouyang, Christina Kim, Christopher Hesse, Shantanu Jain, Vineet Kosaraju, William Saunders, et~al.
\newblock Webgpt: Browser-assisted question-answering with human feedback.
\newblock \emph{arXiv preprint arXiv:2112.09332}, 2021.

\bibitem[OpenAI(2022)]{openai2022chatgpt}
OpenAI.
\newblock Introducing chatgpt.
\newblock 2022.

\bibitem[OpenAI(2023)]{openai2023gpt4}
OpenAI.
\newblock Gpt-4 technical report, 2023.

\bibitem[Rawles et~al.(2023)Rawles, Li, Rodriguez, Riva, and Lillicrap]{rawles2023android}
Christopher Rawles, Alice Li, Daniel Rodriguez, Oriana Riva, and Timothy Lillicrap.
\newblock Android in the wild: A large-scale dataset for android device control.
\newblock \emph{arXiv preprint arXiv:2307.10088}, 2023.

\bibitem[Schuhmann et~al.(2022)Schuhmann, Beaumont, Vencu, Gordon, Wightman, Cherti, Coombes, Katta, Mullis, Wortsman, et~al.]{schuhmann2022laion}
Christoph Schuhmann, Romain Beaumont, Richard Vencu, Cade Gordon, Ross Wightman, Mehdi Cherti, Theo Coombes, Aarush Katta, Clayton Mullis, Mitchell Wortsman, et~al.
\newblock Laion-5b: An open large-scale dataset for training next generation image-text models.
\newblock \emph{Advances in Neural Information Processing Systems}, 35:\penalty0 25278--25294, 2022.

\bibitem[Significant-Gravitas(2023)]{autogpt}
Significant-Gravitas.
\newblock Autogpt.
\newblock \url{https://github.com/Significant-Gravitas/AutoGPT}, 2023.

\bibitem[Singh et~al.(2019)Singh, Natarajan, Shah, Jiang, Chen, Batra, Parikh, and Rohrbach]{singh2019towards}
Amanpreet Singh, Vivek Natarajan, Meet Shah, Yu Jiang, Xinlei Chen, Dhruv Batra, Devi Parikh, and Marcus Rohrbach.
\newblock Towards vqa models that can read.
\newblock In \emph{Proceedings of the IEEE/CVF conference on computer vision and pattern recognition}, pages 8317--8326, 2019.

\bibitem[Sun et~al.(2023{\natexlab{a}})Sun, Fang, Wu, Wang, and Cao]{sun2023eva}
Quan Sun, Yuxin Fang, Ledell Wu, Xinlong Wang, and Yue Cao.
\newblock Eva-clip: Improved training techniques for clip at scale.
\newblock \emph{arXiv preprint arXiv:2303.15389}, 2023{\natexlab{a}}.

\bibitem[Sun et~al.(2023{\natexlab{b}})Sun, Yu, Cui, Zhang, Zhang, Wang, Gao, Liu, Huang, and Wang]{sun2023generative}
Quan Sun, Qiying Yu, Yufeng Cui, Fan Zhang, Xiaosong Zhang, Yueze Wang, Hongcheng Gao, Jingjing Liu, Tiejun Huang, and Xinlong Wang.
\newblock Generative pretraining in multimodality.
\newblock \emph{arXiv preprint arXiv:2307.05222}, 2023{\natexlab{b}}.

\bibitem[Touvron et~al.(2023)Touvron, Martin, Stone, Albert, Almahairi, Babaei, Bashlykov, Batra, Bhargava, Bhosale, et~al.]{touvron2023llama}
Hugo Touvron, Louis Martin, Kevin Stone, Peter Albert, Amjad Almahairi, Yasmine Babaei, Nikolay Bashlykov, Soumya Batra, Prajjwal Bhargava, Shruti Bhosale, et~al.
\newblock Llama 2: Open foundation and fine-tuned chat models.
\newblock \emph{arXiv preprint arXiv:2307.09288}, 2023.

\bibitem[Wang et~al.(2023)Wang, Lv, Yu, Hong, Qi, Wang, Ji, Yang, Zhao, Song, et~al.]{wang2023cogvlm}
Weihan Wang, Qingsong Lv, Wenmeng Yu, Wenyi Hong, Ji Qi, Yan Wang, Junhui Ji, Zhuoyi Yang, Lei Zhao, Xixuan Song, et~al.
\newblock Cogvlm: Visual expert for pretrained language models.
\newblock \emph{arXiv preprint arXiv:2311.03079}, 2023.

\bibitem[Yao et~al.(2022)Yao, Chen, Yang, and Narasimhan]{yao2022webshop}
Shunyu Yao, Howard Chen, John Yang, and Karthik Narasimhan.
\newblock Webshop: Towards scalable real-world web interaction with grounded language agents.
\newblock \emph{Advances in Neural Information Processing Systems}, 35:\penalty0 20744--20757, 2022.

\bibitem[Ye et~al.(2023)Ye, Hu, Xu, Ye, Yan, Xu, Li, Tian, Qian, Zhang, et~al.]{ye2023ureader}
Jiabo Ye, Anwen Hu, Haiyang Xu, Qinghao Ye, Ming Yan, Guohai Xu, Chenliang Li, Junfeng Tian, Qi Qian, Ji Zhang, et~al.
\newblock Ureader: Universal ocr-free visually-situated language understanding with multimodal large language model.
\newblock \emph{arXiv preprint arXiv:2310.05126}, 2023.

\bibitem[Yu et~al.(2023)Yu, Yang, Li, Wang, Lin, Liu, Wang, and Wang]{yu2023mm}
Weihao Yu, Zhengyuan Yang, Linjie Li, Jianfeng Wang, Kevin Lin, Zicheng Liu, Xinchao Wang, and Lijuan Wang.
\newblock Mm-vet: Evaluating large multimodal models for integrated capabilities.
\newblock \emph{arXiv preprint arXiv:2308.02490}, 2023.

\bibitem[Zeng et~al.(2023)Zeng, Liu, Lu, Wang, Liu, Dong, and Tang]{zeng2023agenttuning}
Aohan Zeng, Mingdao Liu, Rui Lu, Bowen Wang, Xiao Liu, Yuxiao Dong, and Jie Tang.
\newblock Agenttuning: Enabling generalized agent abilities for llms.
\newblock abs/2310.12823, 2023.

\bibitem[Zhan and Zhang(2023)]{zhan2023you}
Zhuosheng Zhan and Aston Zhang.
\newblock You only look at screens: Multimodal chain-of-action agents.
\newblock abs/2309.11436, 2023.

\bibitem[Zhu et~al.(2023)Zhu, Chen, Shen, Li, and Elhoseiny]{zhu2023minigpt}
Deyao Zhu, Jun Chen, Xiaoqian Shen, Xiang Li, and Mohamed Elhoseiny.
\newblock Minigpt-4: Enhancing vision-language understanding with advanced large language models.
\newblock \emph{arXiv preprint arXiv:2304.10592}, 2023.

\end{thebibliography}
 }
}%end of hide

\end{document}